\documentclass{article}

\usepackage{arxiv1}

\usepackage{url,hyperref,lineno,microtype,subcaption}

\usepackage{graphicx,soul}
\usepackage{amsmath}
\usepackage{bm}
\usepackage{epstopdf}
\usepackage{epsfig}
\usepackage{subcaption}
\usepackage{caption}
\usepackage{hyperref}
\usepackage{xcolor}
\usepackage{soul}
\usepackage{bbm}
\usepackage{wrapfig}
\usepackage{floatrow}

\usepackage{siunitx}

\usepackage{xr}

\usepackage[utf8]{inputenc} 
\usepackage[T1]{fontenc}    
\usepackage{hyperref}       
\usepackage{url}            
\usepackage{booktabs}       
\usepackage{amsfonts}       
\usepackage{nicefrac}       
\usepackage{microtype}      
\usepackage{lipsum}
\usepackage{graphicx}
\graphicspath{ {./images/} }

\title{SpaRCe: Improved Learning of Reservoir Computing  Systems through Sparse Representations}

\author{
 Luca Manneschi \\
  Department of Computer Science\\
  University of Sheffield\\
  Sheffield, UK, \\
  \small{\texttt{lmanneschi1@sheffield.ac.uk}} \\
   \And
 Andrew C. Lin \\
  Department of Biomedical \\Science, \\ 
  The University of Sheffield, \\ 
  Sheffield, UK,\\
  \small{\texttt{andrew.lin@sheffield.ac.uk}} \\
  \And
Eleni Vasilaki\\
Department of Computer Science\\
  University of Sheffield\\
  Sheffield, UK, \\
   \small{\texttt{e.vasilaki@sheffield.ac.uk}} \\ \\
}

\begin{document}
\maketitle

\begin{abstract}
``Sparse" neural networks, in which relatively few neurons or connections are active, are common in both machine learning and neuroscience. Whereas in machine learning, ``sparsity'' is related to a penalty term that leads to some connecting weights becoming small or zero, in biological brains, sparsity is often created when high spiking thresholds prevent neuronal activity. Here we introduce sparsity into a reservoir computing network via neuron-specific learnable thresholds of activity, allowing neurons with low thresholds to contribute to decision-making but suppressing information from neurons with high thresholds. This approach, which we term ``SpaRCe", optimises the sparsity level of the reservoir without affecting the reservoir dynamics. The read-out weights and the thresholds are learned by an on-line gradient rule that minimises an error function on the outputs of the network. Threshold learning occurs by the balance of two opposing forces: reducing inter-neuronal correlations in the reservoir by deactivating redundant neurons, while increasing the activity of neurons participating in correct decisions. We test SpaRCe on classification problems and find that threshold learning improves performance compared to standard reservoir computing. SpaRCe alleviates the problem of catastrophic forgetting, a problem most evident in standard echo state networks and recurrent neural networks in general, due to increasing the number of task-specialised neurons that are included in the network decisions.
\end{abstract}

\section{Introduction}
%
%
%
%
The performance of artificial neural networks is often improved by adopting ``sparse" representations, in which relatively few neurons or connections are active.
Previous research has studied the role of sparse connectivity, in terms of memory, in Hopfield networks,  demonstrating how sparse connectivity increases storage capacity \cite{tsodyks1988enhanced} \cite{tsodyks1988associative} \cite{derrida1987exactly} \cite{amit1985storing}.
 Memory retrieval and associative learning have been studied in the context of neural network attractors, and the work in \cite{romani2013scaling} has provided an abstract mathematical analysis of retrieval capacity. 
From the machine learning perspective, adopting sparse connectivity can lead to more interpretable models \cite{hastie2015statistical}, a reduced computational cost \cite{NIPS2016_6504}, and can help solve overfitting problems \cite{srivastava2014dropout}. 
sparsity in machine learning is typically introduced to artificial networks through regularisation, in which a penalty term leads to the reduction of the connection weights.
In this regard, the work in \cite{NIPS2016_6504} demonstrated how structured sparsity can improve computational speed and accuracy in a convolutional neural network. Rasmussen et al.\ \cite{rasmussen2012model} showed how the choice of regularization parameters of the model can impact the interpretability and the reproducibility of a classifier of neuroimaging data, and showed the existence of a trade-off between pure classification accuracy and reproducibility. 

Sparsity is also a well-known concept in neuroscience: biological neurons are highly selective in systems ranging from mammalian sensory cortex  \cite{rolls1995sparseness} to the insect mushroom body \cite{honegger2011cellular}  \cite{lin2014sparse}. However, unlike in typical machine learning approaches, biological sparsity is introduced not only by reducing connection weights between neurons, but also by the fact that neurons have spiking thresholds: they only fire when their summed inputs exceed a certain threshold. High spiking thresholds relative to the size of synaptic inputs can often contribute to high selectivity of neurons, as with Kenyon cells (KCs), the principal neurons of the insect mushroom body, which fire sparsely in response to odor stimuli \cite{turner2008olfactory} \cite{gruntman2013integration} \cite{li2013transformation} \cite{perez2002oscillations}. In the fruit fly \emph{Drosophila}, this sparse odor coding enhances learned discrimination of similar odors \cite{lin2014sparse}. Moreover, spiking thresholds vary across neurons \cite{turner2008olfactory} and over time for the same neuron \cite{jeanne2015convergence} \cite{azouz2000dynamic}, and spiking thresholds for different neurons are adapted to neurons' particular input statistics \cite{jeanne2015convergence} and past activity \cite{grubb2010activity}.
Here we applied the concept of adaptable spiking thresholds to machine learning to create SpaRCe, a method for achieving sparse representations applicable to Reservoir Computing.

Reservoir computing takes inspiration from the complex non-linear behavior of recurrent neural networks and their abilities to process temporal information. Such a computing paradigm exploits the inherent complexity of a dynamical system, which is called the reservoir and that can be virtually or physically defined, to represent a stream of data into a high dimensional space useful for learning. In the classical learning framework of such systems, training occurs on the read-out connections only, which connect the activities of the non-linear reservoir components to the output neurons. The advantages of the reservoir computing framework are: first, the system can be physically defined as in \cite{du2017reservoir,kulkarni2012memristor,zhu2020memristor,vandoorne2014experimental,paquot2012optoelectronic}, aspect that can dramatically reduce the computational and energetic cost \cite{nakajima2020physical}; second, the response of the reservoir can exhibit memory over a range of timescales without training of the internal dynamic of the system. In the latter aspect, the reservoir computing framework is antithetical to standard recurrent neural networks, whose connectivity is trained through the computationally expensive BPTT algorithm. 

In this work, we model the reservoir as a recurrent network of leaky integrators \cite{jaeger2007optimization} known as echo state network (ESN). The connectivity between the nodes is represented through a random sparse fixed adjacency matrix, whose spectrum of eigenvalues allows the system to exhibit a wide range of timescales and to efficiently represent temporal information. 
Previous works have explored alternatives to the typical Erdos-Renyi connectivity of the reservoir, imposing a regular structure as a delay line \cite{rodan2010minimum}, where the nodes are connected unidirectionally composing a circular graph, or defining the graph connectivity through more complex topologies \cite{cui2012architecture} as scale-free or small world. Furthermore, complex structures composed by multiple, interconnected ESNs have been recently studied, and the works \cite{gallicchio2017deep} \cite{manneschi2020} demonstrate how the use of different hyperparameters for different ESNs can expand the range of timescales that is accessible by the system and consequently increase the quality of the reservoir representation for temporal tasks with complex temporal dynamic. In this regard, the work \cite{manneschi2020} gives insights into the understanding of hierarchical systems from the point of view of accessible timescales, offering a theoretical and a practical analysis on how to tune the hyperparameters of the system to a considered task.

While ESNs are most traditionally adopted by the machine learning community for time-series prediction and forecasting \cite{jaeger2007optimization}, more recent works have studied their behavior for time-series classification \cite{bianchi2020reservoir}. In particular, \cite{bianchi2020reservoir} studies and compares different methods to exploit the representation of the reservoir as input to more complex machine learning algorithms as multilayer perceptrons or SVM.
In this work, we formulate a different approach to improve the classification ability of reservoir systems in general while maintaining the low cost of computation that is intrinsic to the concept of reservoir computing. 
Analogously to the concept of firing thresholds, SpaRCe exploits learnable thresholds to optimize the level of sparsity inside the network. Both the learnable thresholds and the read-out weights (but not the recurrent connections within the reservoir) are optimised by minimising an error function that does not include any normalization term. We analysed the learning rule derived from this error minimisation and found that learning occurs by two antagonist factors: the first raises the thresholds proportionally to the correlated activity of the nodes (thus silencing nodes that are correlated and therefore redundant), while the second lowers the thresholds of nodes that contribute to the correct classification (Fig.\ \ref{two_forces}). The novelty of our work lies in the fact that a sparsity level is reached due to the presence of ``on-line'' learnable firing thresholds, rather than to penalty terms \cite{huang2011learning} \cite{hastie2015statistical} \cite{candes2008enhancing}, as well as in the detailed analysis on how such thresholds enable ESNs to perform competitively in tasks where their performance was lacking. 

\hfill 
 
\hfill 

\section{Methods}

\subsection{Standard Echo State Network \label{ESN_intro}}

An echo state network (ESN) is composed of randomly connected leaky integrators. 
The activity of such a system is defined through the following equation \cite{jaeger2007optimization}:
\begin{equation}
\textbf{V}(t+\delta t)=(1-\alpha)\textbf{V}(t)+\alpha f\big[\gamma \mathbf{W}_{in}\textbf{s}(t)+\rho \mathbf{W}\textbf{V}(t)\big]
\label{V}
\end{equation}
where $\alpha=\frac{\delta t}{\tau}$ defines the temporal scale of the neuron, $\textbf{V}(t)$ is the activity vector of the integrators and $\textbf{s}(t)$ is the input signal.
$\mathbf{W}$ is the fixed sparse random matrix that describes the recurrency of the reservoir. The matrix $\mathbf{W}$ is random and sparse, corresponding to an Erdos-Renyi graph where the probability of a connection is $p_{ER}$. The eigenvalues of $\mathbf{W}$ are rescaled to be confined inside the unit circle of the imaginary plane, necessary condition for the echo state property. The hyperparameter $\rho$ (between 0 and 1) controls further the radius of the spectrum of the system's eigenvalues. Finally, $\gamma$ is a gain factor of the input signal. The specific form of the input matrix $\mathbf{W}_{in}$ and the activation function $f$ is task-dependent and will be specified in sections \ref{task_bio} and \ref{task_ML}.\

It is possible to control \textit{a priori} the range of timescales that the reservoir exhibits by appropriately choosing $\alpha$ and $\rho$ as described in the methodology reported in Appendix A.1 and our earlier work \cite{manneschi2020}. In the standard ESN learning protocol, learning occurs exclusively on the read-out defined from a vector $\tilde{\textbf{V}}$  that represents an ensemble of the reservoir activities $\textbf{V}$.
The output of the system is then computed through 
\begin{align}
\textbf{y}=\textbf{W}_{o}^{T} \tilde{\textbf{V}} \\
E=\dfrac{1}{2}  \Big[\tilde{\textbf{y}}-\textbf{y}\Big]^2
\end{align}
where $\tilde{\textbf{y}}$ denotes the desired output values, and the output connectivity $\textbf{W}_{o}^{T}$ is optimised to minimise the cost function $E$ through ridge regression \cite{jaeger2007optimization}  or through gradient descent methods \cite{lukovsevivcius2012practical}. 
We note how, depending on the task and on the learning framework, the $\tilde{\textbf{V}}$ vector can be defined from different ensembles of activities of the reservoir across time. 
The idea to exploit the dynamic of the ESNs, instead of using its activities at one step only, is particularly useful for classification tasks. Previous works have introduced different techniques to define a $\tilde{\mathbf{V}}$ vector from the ESN's dynamic: adopting PCA on the multidimensional response of the reservoir across time \cite{bianchi2020reservoir}, introducing a learnable temporal kernel to define the read-out \cite{ma2016functional}, or concatenating the past activities of the reservoir to define a higher dimensional vector $\tilde{\mathbf{V}}$ \cite{lukovsevivcius2012practical} \cite{schaetti2016echo}. In this work, we will exploit the latter approach, while other techniques such as PCA, which reduces the dimensionality of the ESN's representation, are more desirable in the case of overfitting. In this regard, while the most spontaneous choice for time-series prediction would be $\tilde{\textbf{V}}=\textbf{V}(t)$, corresponding to the activity of the reservoir at the current time, it is also possible to expand the dimensionality of the system by including previous reservoir representations at times $t_l$ and define a vector $\tilde{\textbf{V}}=\mathcal{C}\Big( \big\{\textbf{V}(t_l)\big\}_{t_l \varepsilon \mathcal{T}} \Big)$, where $\mathcal{C}$ denotes the concatenation operator and $\mathcal{T}$ the ensemble of time steps $t_l$ considered.  This latter approach has been adopted also in physical reservoir models through the use of virtual nodes \cite{torrejon2017neuromorphic} \cite{tanaka2019recent}. Of course, such dimensionality expansion leads to an artificial increase of the memory of the system. However, the concatenation of previous activities does not guarantee the understanding of the dependencies among events that are distant in time, because the memory of a past input signal could have faded away when a new stimulus $\textbf{s}(t)$ comes. A scheme of this procedure for time-series classification can be found in Fig.~\ref{Figure1}~A.

\subsection{Sparse Reservoir Computing \label{SpaRCe}}

\begin{figure*}[h!]
 \centering
    \makebox[\textwidth][c]{\includegraphics[width=1.\textwidth]{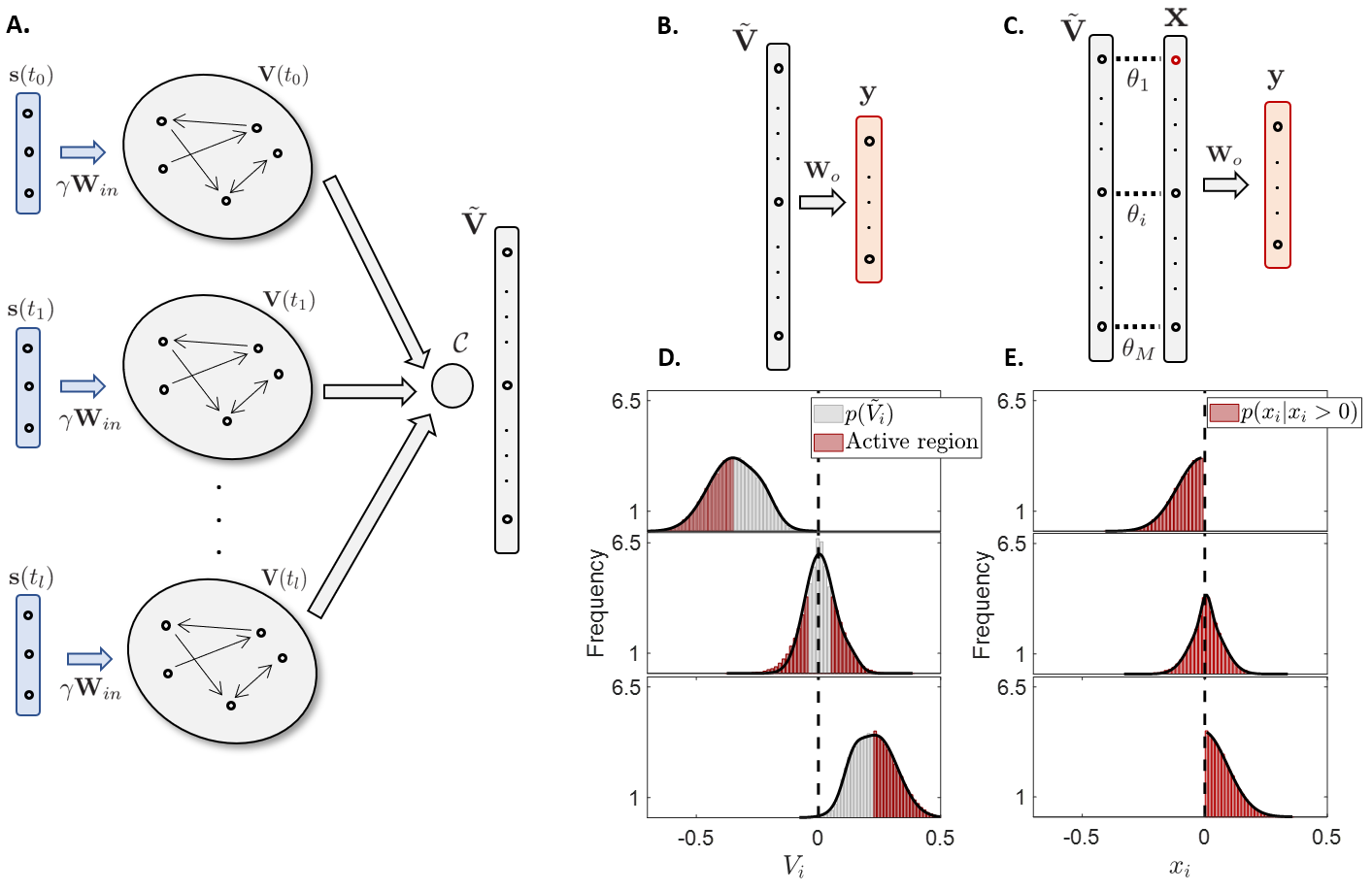}}
    \caption{\small{Comparison between the basic reservoir computing framework and SpaRCe, which exploits the concepts of adaptable thresholds to introduce sparsity in the representation of the ESN. \textbf{A}: The typical reservoir computing sampling paradigm for a time-series classification task. Time flows from top to bottom. The input signal $\textbf{s}(t)$ is fed into the ESN through the input connectivity matrix $\gamma \textbf{W}_{in}$. An ensemble of the activities $\textbf{V}(t)$ across time is selected and concatenated to compose a vector $\tilde{\textbf{V}}$, which is then used to define the read-out. Typical choices of ESN activities used to define the read-out are $\tilde{\textbf{V}}=\textbf{V}(T)$, where $T$ denotes the final temporal step of the input sequence, or $\tilde{\textbf{V}}=\mathcal{C}\Big(\big\{\textbf{V}(t)\big\}_{\forall t}\Big)$, denoting the concatenation of all the reservoir dynamic across the temporal length of $\textbf{s}(t)$. Of course, while the first approach relies on the ESN intrinsic memory capacity, the latter case corresponds to a dimensionality expansion that contributes artificially to the memory of the system. Both these approaches and intermediate cases, i.e. where the dynamic of the reservoir across time is sampled with low frequency to compose the read-out, will be exploited in the tasks studied.   \textbf{B-C}: Comparison between the typical reservoir computing read-out (\textbf{B}) and the SpaRCe model (\textbf{C}).  \textbf{B}: The ESN output $\textbf{y}=\textbf{W}_o \tilde{\textbf{V}}$ is responsible for the classification process of the example sequence $\textbf{s}(t)$. In this paradigm, learning occurs exclusively on $\textbf{W}_o$. \textbf{C}: Scheme of the SpaRCe model. Thresholds are introduced at the level of the $\tilde{\textbf{V}}$ vector, leaving unaffected the dynamic of the reservoir and making the approach applicable to any physical or virtual reservoir model. Each threshold value is composed by a normalisation term $\textbf{P}_n(|\tilde{\textbf{V}}_i|)$, defined as the n-th percentile of the activity distribution of the i-th component across the data, plus an adaptable term  $\tilde{\theta}_i$ (see text for more details). \textbf{D-E}: Activity distributions of $\tilde{\textbf{V}}$ (\textbf{D}) and \textbf{x} (\textbf{E}) for three example nodes. The highlighted red region in \textbf{D} corresponds to the values for which the nodes would be active if the normalisation mechanism proposed in Eq.~\ref{x} would be applied  (percentile $n=50$ in this case). From \textbf{E}, it is clear that Eq.\ref{x} also shifts the activity distributions acting as a normalisation mechanism.   }}
  \label{Figure1}
\end{figure*}

\begin{figure*}[h!]
 \centering
  \includegraphics[width=0.85\textwidth]{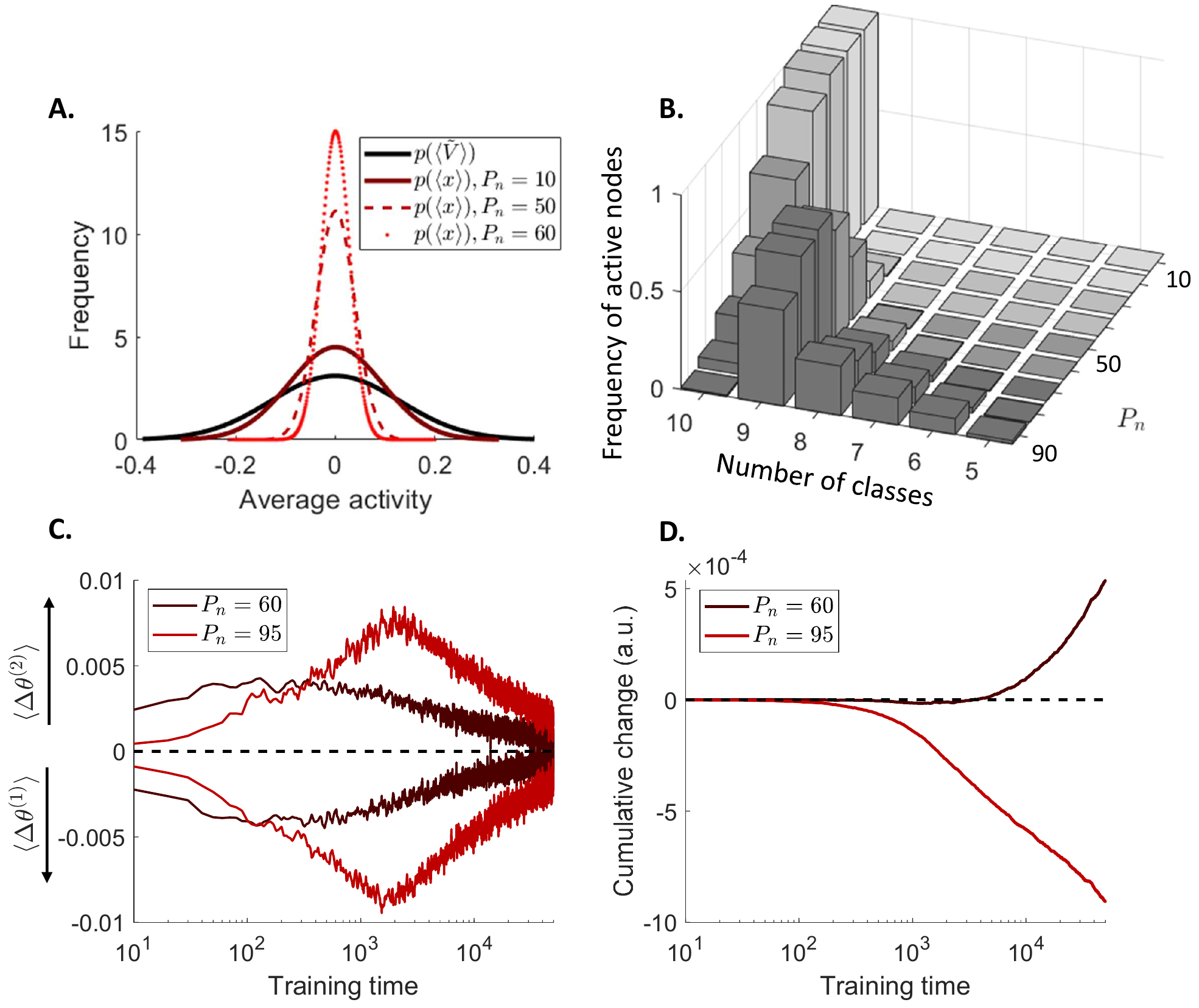}
    \caption{\small{ The SpaRCe read-out proposed leads to decreased variability in the ESN representation, to specialised responses, and to a interpretable learning rule that acts as feature selection mechanism. The learning rule for the thresholds is driven by the imbalance between two antagonist forces. \textbf{A} Distributions of the average activities of nodes across an example dataset (MNIST, section \ref{Benchmarks}) for the standard ESN and for SpaRCe as the percentile of the normalisation mechanism changes at the beginning of training (when $\tilde{\boldsymbol{\theta}}=0$). SpaRCe descreases the variability of the activity distributions, acting as a normalisation mechanism. 
    \textbf{B}: Frequency of active nodes for different starting sparsity levels and different number of classes for a classification task with ten classes (MNIST, section \ref{Benchmarks}) before training ($\boldsymbol{\tilde{\theta}}=0$). As sparsity ($P_n$) increases, nodes respond to a smaller number of classes becoming more specialised.
    \textbf{C}: Analysis of the two forces $\Delta \theta^{(1)}$ and $\Delta \theta^{(2)}$ involved in the learning rule for the thresholds. The positive y-axis shows a running average of $\Delta \theta^{(1)}$ with solid lines, while the negative y-axis shows a running average of $\Delta \theta^{(2)}$ with dashed lines. $\langle \rangle$ indicates averaging across all neurons. $\Delta \theta^{(1)}$ increases the threshold values for nodes that are equally contributing to the classification process. $\Delta \theta^{(2)}$ decreases the threshold values thanks to the positive contribution of the output weights connected to the correct output. Colours correspond to initial conditions ($P_{10}, P_{60}$).  \textbf{D}: Average cumulative change of a threshold. If the starting level of sparsity is suboptimal and low (high) the average threshold change is positive (negative).  }}
  \label{two_forces}
\end{figure*}

In contrast to previous models that define the output of the neural network through a read-out of the $\tilde{\textbf{V}}$ vector, i.e. the activities of the reservoir considered for the learning process,  we introduced another variable $x_i$ for each dimension of $\tilde{\textbf{V}}$, defined as follows:

\begin{align}
x_i=sign\Big(\tilde{V}_i\Big)relu\Big\{ |\tilde{V}_i|-\theta_i \Big\} \label{x} \\
\theta_i=P_{n}\Big(|\tilde{\textbf{V}}_i|\Big)+\tilde{\theta}_i \label{theta_i}
\end{align}
where $relu$ stands for rectified linear unit,  $sign$ is the sign function (1 if $\tilde{V}_i>0$, -1 if $\tilde{V}_i<0$, 0 if $\tilde{V}_i=0$), and $\theta_i$ is a threshold that enables $x_i$ to be sparse. Thus, the variable $x_i$ is zero if the absolute value of the variable $V_i$ is lower than the corresponding threshold. We term this variant SpaRCe for Sparse Reservoir Computing. Of course, $\textbf{x}=\tilde{\textbf{V}}$ if $\bm{\theta}=0$, which is the case where the formulated learning procedure coincides with the standard ESN paradigm. Considering Eq.~\ref{theta_i}, each threshold is composed by two factors:
\begin{itemize}
\item $P_{n}(|\tilde{\textbf{V}}_i|)$, defined as the percentile $n$ of the distribution of activity of the i-th component of $|\tilde{\textbf{V}}|$ across a dataset, where the same value of $n$ is applied to all dimensions of $\tilde{\textbf{V}}$. Given that the response $\tilde{\textbf{V}}$ of the reservoir to a specific input remains unchanged across learning, this term can be computed over the training dataset and maintained constant throughout the simulation. Thus, the role of $P_{n}(\tilde{|\textbf{V}}_i|)$ is to initialise the thresholds from an initial condition where the sparsity level of all dimensions of $\textbf{x}$ is $n/100$. The thresholding mechanism introduced is symmetric with respect to zero, as is visible from the examples in Fig.\ \ref{Figure1}D-E. While panel D highlights the portions of the distributions that are positive after normalisation, panel E shows the resulting distributions after application of Eq.\ \ref{theta_i}. In this formulation, the value of $n$ needs to be considered as an additional, interpretable hyperparameter. However, we will show that all the sensible choices of $n$  will correspond to faster convergence and better performance of the formulated algorithm in comparison to the standard ESN learning paradigm across all tasks considered. Throughout this paper, the same $P_{n}(\tilde{|\textbf{V}}_i|)$s computed over the training dataset will be used for the validation and test dataset.  \newline

\item An adaptable component $\tilde{\theta}_i$ optimised through gradient descent learning rules. This parameter adapts the sparsity level for the considered task and can rediscover the standard learning paradigm in the case where $\tilde{\theta}_i=-P_{n}(|\tilde{\textbf{V}}_i|)$, $\forall i$. 
\end{itemize}

The additional complexity arising from Eq.\ \ref{x} is summarized by Fig.~\ref{Figure1}~B, which depicts the difference between the read-out of a standard Echo-State network and our formulation. 
Eq.~\ref{x} acts as normalisation operator that directly controls sparsity in the reservoir representation and that, thanks to the learnable components $\tilde{\boldsymbol{\theta}}$, works as a feature-selection mechanism. In the latter sense, we can now demonstrate the interpretability of the updating equations on $\boldsymbol{\theta}$ derived from a gradient descent approach. 
For the case of a mean square error function, the learning rule can be factorised in two terms:
\begin{align}
\Delta \theta_k=\eta_{\theta}\Big[ \Delta \theta^{(1)}_{k}+\Delta \theta^{(2)}_{k}\Big]\\
\Delta \theta^{(1)}_{k}= \sum_{j=1}^{N_{class}}y_jW^{o}_{jk} sign\big(x_k\big)= \nonumber\\ = \sum_{j=1}^{N_{class}}\sum_{l=1}^{N}W^{o}_{jl}W^{o}_{jk}x_l sign\big(x_k\big) \label{theta_+_main}\\
\Delta \theta^{(2)}_{k}=- W^{o}_{\tilde{j}k}sign\big(x_k\big) \label{theta_-_main}
\end{align}
where $\eta_{\theta}$ is the learning rate on the thresholds,  $\tilde{j}$ refers to the correct output class for the sample considered, $W^{o}$ is the output connectivity matrix. Eq.\ \ref{theta_+_main} and \ref{theta_-_main} are derived by assuming one-hot encoding (Appendix B.2).
Considering $x_k>0$ (analogous considerations are valid for $x_k<0$), the factor $\Delta \theta^{(2)}$ decreases (increases) the threshold value of nodes with $ W^{o}_{\tilde{j}k}>0$ ($W^{o}_{\tilde{j}k}<0$) that help the network reach the right (wrong) classification. 
Thus, $\Delta \theta^{(2)}$ is driven by the output weight between the considered node (if it is active) and the desired class. In contrast, $\Delta \theta^{(1)}$ is a measure of correlation of activities between different nodes in the reservoir and increases the thresholds of neurons that have coherent synapses and that are simultaneously active (and are therefore redundant).  

We examined the two factors $\Delta \theta^{(1)}$ and $\Delta \theta^{(2)}$ across learning for an example of sequence classification and for different initial sparsity levels.  These are plotted in Fig.\ \ref{two_forces}C, with $\Delta \theta^{(1)}$ in solid lines on the positive y-axis and $\Delta \theta^{(2)}$ in dotted lines on the negative y-axis. The two forces are almost symmetric, but their slight imbalance provides the direction to change the threshold values. Indeed, if the starting sparsity level is high, the total force is negative and the factor $\Delta \theta^{(2)}$ dominates, while if the sparsity level is low, the correlation term $\Delta \theta^{(1)}$ wins and the thresholds increase on average (compare $P_{70}$ and $P_{10}$ in Fig.\ \ref{two_forces}D, which shows cumulative $\Delta \theta^{(2)}$ and $\Delta \theta^{(1)}$). The reason that the magnitudes of the forces are larger for a higher starting sparsity (Fig.\ \ref{two_forces}C) is that stimulus representations overlap less when the sparsity level is higher and neurons are more specialised, i.e. preferably fire for one pattern over the others. This leads to more coherent sign of the output weights of the nodes belonging to a cluster toward a specific class, which increases $\Delta \theta^{(1)}$ according to Eq.\ \ref{theta_+_main}. 
A similar analysis of the learning rule for a cross entropy cost function is reported in Appendix B.2.
 Fig.\ \ref{two_forces}A,B describes the benefits due to the application of the proposed normalisation mechanism: first, it directly controls sparsity and introduces specialised neurons (Fig.\ \ref{two_forces}B); second, it shifts the activity distributions $\tilde{\textbf{V}}$ of the network into a more stereotyped response $\textbf{x}$, avoiding the possibility that learning could be dominated by the nodes with highest activities (Fig.\ \ref{two_forces}A). Finally, it stabilises the learning process for a wide range of possible learning rates that would not be accessible without thresholds (Supplementary Materials).
This specific formulation allows the model to use local information to learn the threshold values and optimize sparse representations. We note also how Eq.\ \ref{x} does not affect the timescales of the network and consequently preserves the idea of reservoir computing as a fixed, dynamically rich representation. Furthermore, the method formulated constitutes a computationally inexpensive procedure that is easily applicable to any type of reservoir, virtually or physically defined. \newline \newline

\subsection{Benchmarks \label{Benchmarks}}

We tested the proposed model on the following benchmarks, comparing its performance to the standard read-out of an ESN:

\begin{itemize}
\item A biologically inspired task to test the storage capacity of the system in the memorisation of associations between sequences and desired output values.  In this case, the model is tested on the same associations used for training, but with different realisations of multiplicative white noise added on the sequences. The dependence of SpaRCe on the starting sparsity level is analysed along comparisons of the performance achieved with other methods for reading out from the ESN representation, including deep feedforward networks. \newline
\item Three variants of classification tasks based on the MNIST dataset, where data are presented to the system as temporal varying sequences. In the first task, we studied the performance of the model when each image, composed by $28\times28$ pixels, is presented column by column as a 28-dimensional sequence of 28 time steps (MNIST \cite{schaetti2016echo}). In the second, the input is processed in the same way, but a random permutation of the pixels is applied to all data to make the task more challenging and to randomise the structures of the images (pMNIST). In the third, we applied a random permutation of the data as in the second paradigm, but each image is presented pixel by pixel as a one dimensional sequence of 784 temporal steps (psMNIST \cite{chandar2019towards}).  When the initial conditions of the model are studied, the performance are computed and shown on the test set as $P_n$ varies for comparison, but to select the best $P_n$ value and report the highest performance we used a validation dataset. Of course, we selected the hyperparameters corresponding to the highest accuracy on the validation set and then computed the performance on the test dataset.  \newline   
\item Two tasks that involve sequential learning. In the first case, we applied ten different permutations to the data and trained the model processing the tasks one after the other \cite{kemker2018measuring} \cite{serra2018overcoming}. In the second, the network is trained on the data belonging to different classes sequentially \cite{kemker2018measuring} \cite{serra2018overcoming}. In both cases, the network can access a specific dataset (corresponding to a permutation or a class) only once during training.  As before, a validation dataset is adopted when selecting the best hyperparameters (in particular $P_n$) of the proposed model.
\end{itemize}

\section{Results}

\subsection{Threshold learning increases storage capacity} \label{task_bio} 

We evaluated the performance of a standard ESN and SpaRCe in classifying an ensemble of sequences $\{\textbf{S}_i\}_{i=1,..,N_{seq}}$ of three successive stimuli, where the dimensionality of the signal is $N_{In}=24$. Each stimulus of a sequence is derived from the simulated response of $N_{In}=24$ projection neurons (PNs, in the fly olfactory system) to 110 different odors \cite{hallem2006coding} \cite{olsen2010divisive}. This simulated activity, which we call $\textbf{s}^{HO}$ (HO for Hallem-Olsen), has previously been used in computational analyses of fly olfaction \cite{luo2010generating} \cite{parnas2013odor} \cite{krishnamurthy2017disorder}. 

 Sequences are generated to test the storage capacity of the system in memorising associations between input signals and desired output values. The procedure for building different sequences from single stimuli is described the Appendix (section C.1, Fig. 1), but essentially, it guarantees that each sequence has to be classified as independently as possible from other sequences, and that there are no correlations of elements among different inputs that can inform the classification process. Thus, the system can only memorise the associations among a specific succession of elements and the corresponding output value. 

There are three stimuli in a sequence, each of them is presented for a time interval $\Delta t=0.1 s$ 
in order to allow the network to integrate the information. The total duration of an input sequence is $T=0.3 s$. 
Given a sequence $\textbf{s}_i(t)$, we added multiplicative white noise to each separate dimension to make the task more complex. Thus, the i-th dimension of the final sequence $\textbf{S}_i(t)$ to be classified is $S_i(t)=s_{i}(t)+\sigma_s\xi(t)s_i(t)$, where $\xi(t)$ is a Gaussian distributed random variable with zero mean and unitary variance. 

For this specific task, the activation function $f$ of Eq.\ \ref{V} is a rectified linear unit and the connections of the input adjacency matrix $\textbf{W}_{in}$ follow a lognormal distribution. This particular form of $\textbf{W}_{in}$ is inspired by the biological results in \cite{turner2008olfactory} \cite{caron2013random} \cite{song2005highly}. 
In this case, $\tilde{\textbf{V}}=\textbf{V}(T)$, meaning that only the activities at the last temporal step of a sequence are adopted for the read-out, and the memory of past events is left to the internal dynamic of the ESN. A schematic representation of the task can be found in the Appendix.

\begin{figure*}[h!]
 \centering
    \makebox[\textwidth][c]{\includegraphics[width=1.0\textwidth]{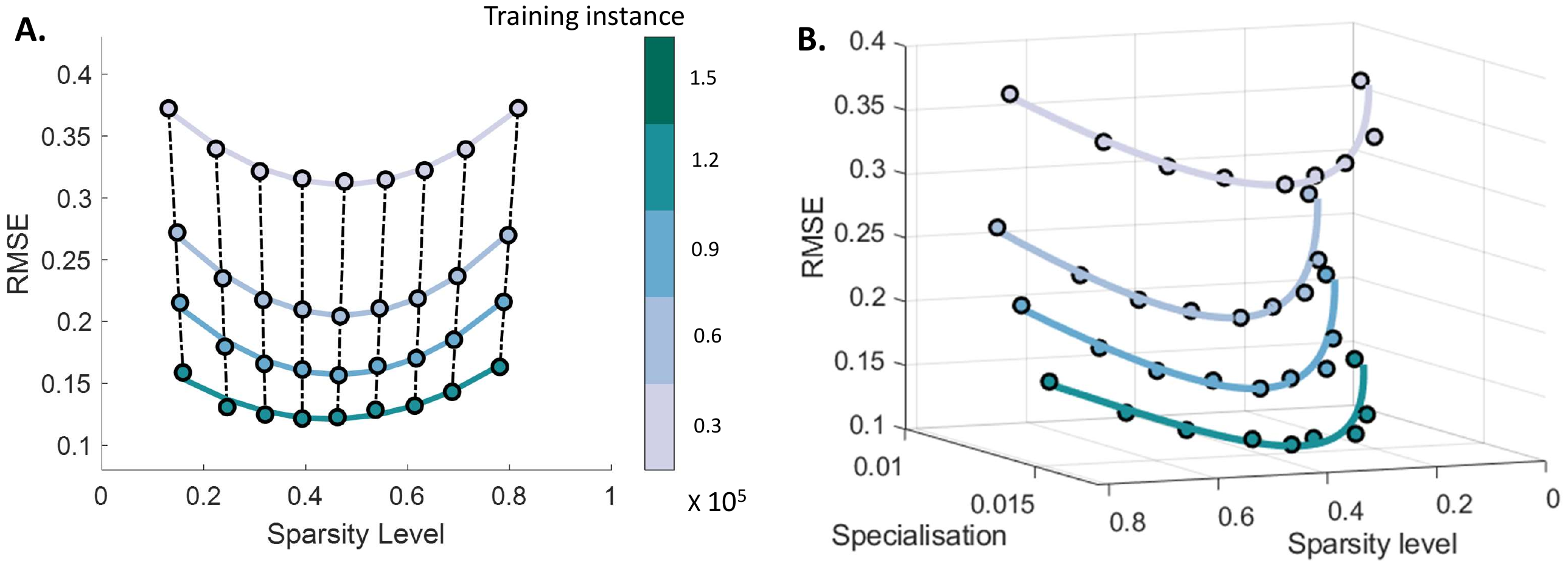}}
    \caption{\small{The learning process modulates the network's sparsity level toward an optimal percentage of active nodes. \textbf{A}: Performance as a function of sparsity for different training instances of the model (a color represents a specific training time, which increases from top to bottom). For each instance the results are fitted with a second degree polynomial that has a minimum around $0.5$ on the x-axis, demonstrating the existence of an optimal percentage of active nodes. The dashed line connecting the results for diverse training time highlights the change in the sparsity level achieved through the learning rule. \textbf{B}: Performance as a function of sparsity and specialisation. The best performance corresponds to the highest specialisation values for all training instances, demonstrating the interpretability of the model.}}
  \label{Figure2}
\end{figure*}

We first investigated how the model depends on initial sparsity, and found an optimal sparsity level for the task. We initialized the network with different initial sparsity percentiles (i.e., thresholds at different percentiles of the $\textbf{V}$ distribution, $n=[10,20,30,40,50,60,70,80,90]$), and tracked the mean square error over the learning process (Fig.\ \ref{Figure2}A). Errors decreased as learning proceeded, but at each time point, the lowest error occurred for an initial sparsity of about $50\%$. Furthermore, models initialized with sparsity values other than $50\%$ (an explanation of why $50\%$ can be found in the Appendix) converged toward $50\%$ as training proceeded, as shown by the black dashed lines connecting dots of training instances from the top to the bottom of the graph (Fig.\ \ref{Figure2}A). This shows how the learning rule pushes the percentage of active nodes toward the optimal sparsity level. 

Notably, the error is smallest when specialisation is highest (for a definition of specialisation, see Appendix A.2, Eq. A5; conceptually, specialisation represents to what extent a node is active for stimuli of one class, not stimuli of other classes), as shown in Fig. \ref{Figure2}B, which reports the error as a function of sparsity and specialisation. For all training instances analysed, the lowest error corresponds to the highest specialisation value. Thus, specialisation provides a systematic way to choose the starting condition of the network. Indeed, it is possible to select the thresholds using the percentile value that yields the highest specialisation measure. However, there is no need to excessively fine-tune the initialization, since the learning rule will optimize the threshold values anyway. We note also that this simulation is performed through a simple gradient descent algorithm, and that the model's dependence on initial conditions can be ameliorated by using more complex optimizers, as will be shown in section \ref{task_ML}. 

\begin{figure*}[h!]
 \centering
    \makebox[\textwidth][c]{\includegraphics[width=1.\textwidth]{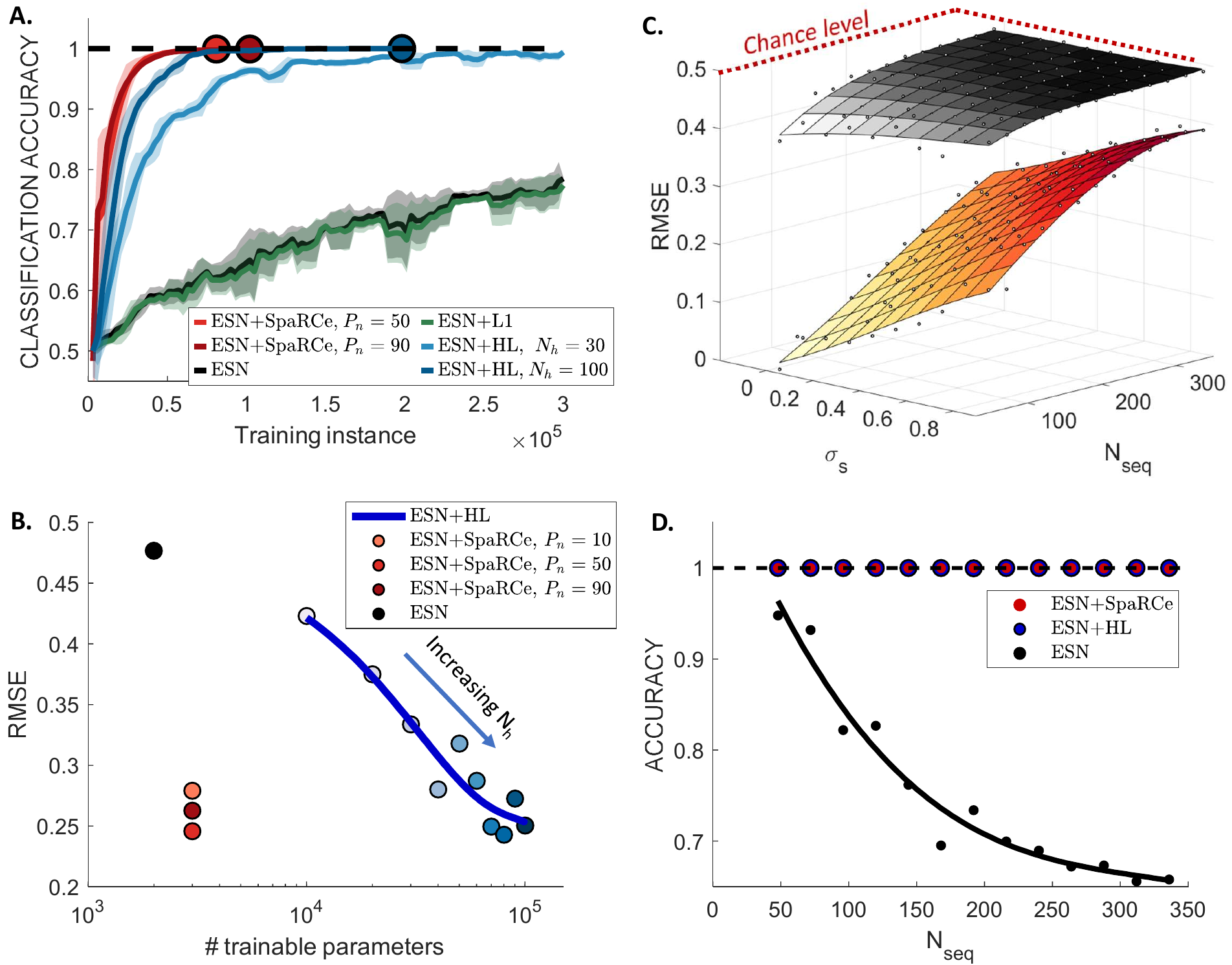}}
    \caption{\small{The SpaRCe algorithm increases the memory capacity of the ESN and the stability of the found solution. \textbf{A}: Classification accuracy and root mean square error of the models for a case where the number of sequences to be classified is $192$. Each minibatch corresponds to the presentation of $20$ training samples. In this case, the performance of SpaRCe are related to a starting $P_n=50$, while the hidden layer has $N_h=100$ nodes. Dots correspond to the training instances in which the considered models solve the task (for the models that can solve the task), thus showing the training speed of the algorithms analysed. \textbf{B}: Performance of SpaRCe for three different starting sparsity levels (red, orange colours), for a standard ESN (black) and of HL (blue colours) for diverse $N_h$ nodes. The x-axis reports the number of trainable parameters and is shown in a logarithmic scale. The graph shows how SpaRCe is able to reach good performance while maintaining a low number of trainable parameters. \textbf{C}: Comparison of the root mean square error for the ESN and the ESN with thresholds as the external noise $\sigma_s$ and the number of stimuli vary. The introduction of thresholds lead to robust result. \textbf{D}: Performance as the number of inputs to be classified increases, for the hidden layer model with $N_h=100$ (blue),  SpaRCe (red) with $P_n=50$ and a standard ESN (black). SpaRCe and HL solves the tasks considered, but the latter uses a number of trainable parameters (reported along the performance with the colour scheme that reflects the referred model) that is of two magnitudes higher than the first. The accuracy of the standard ESN drops considerably as the number of sequences increases. The inset shows the root mean square error for SpaRCe and for HL, varying the starting condition $P_n$ and number of nodes $N_h$ respectively. The errors shown correspond to the training instance in which the fastest model reaches perfect classification accuracy. Numbers reflect the number of trainable parameters for example cases.}}
  \label{Performance}
\end{figure*}

Finally, we compared the performance of the SpaRCe model with:
\begin{itemize}
\item[1)] Echo state network (ESN) without thresholds, where the same on-line learning is applied to the output weights $W^{out}$ only. We note that the algorithm SpaRCe learns $N$ more parameters (the thresholds) in comparison to the Echo State Network without thresholds. 
\item[2)] Hidden layer (HL), where we added a full hidden layer of $N_h$ nodes on the top of the reservoir. This approach learns an additional connectivity matrix between the reservoir and the hidden layer, dramatically increasing the number of parameters by a factor of approximately $N_h N$. 
\item[3)] Echo state network (ESN) with online learning and \textit{$L_1$} or \textit{$L_2$} regularization terms on the output weights. 
\end{itemize}  
The SpaRCe model outperforms the standard ESN with or without the regularization terms, based on classification accuracy and root mean square error (Fig.\ \ref{Performance}A,C,D). This advantage is consistent across different levels of external noise ($\sigma_s$) and different numbers of stimuli (Fig.\ \ref{Performance}C,D). Furthermore, SpaRCe performs comparably to a network with an additional full hidden layer with $N_h\approx100$ nodes, even though the hidden layer dramatically increases the number of learnable parameters compared to SpaRCe (Fig.\ \ref{Performance}A,B,D). In comparison to the addition of a hidden layer, the model SpaRCe provides a cheap formulation to achieve an optimal and reliable sparsity level (see Fig.\ \ref{Performance}B, where the number of learnable parameters for the models are reported with the corresponding performance). 

In general, when introducing a hidden layer of $N_h$ neurons trained with backpropagation, for a network of $N_o$ output neurons, we would learn additional parameters $N \times N_h+N_h+N_h \times N_o + N_o  = N_h\times (N+N_o+1)+ N_o$ (weights + biases + output weights + biases, $1000\times100+100+100\times2+2\approx 10^{5}$ for the case with $N_h=100$), while for a classical reservoir we learn $N \times N_o+ N_o \approx N \times N_o$ ($1000\times2+2\approx 2\times10^{3}$) and for SpaRCe $N \times N_o+ N + N_o\approx  (N+1)  \times N_o $ ($1000\times2+1000+2\approx3\times10^3$). 
While adding a hidden layer goes against the principle of ESN of exploiting the network dynamics while using simple learning methods, it remains an interesting comparison for quantifying the efficiency of the proposed method. With  $N_h=100$ and while $N_o<<N_h$ our proposed thresholded architecture is efficient in terms of numbers of learnable parameters in comparison to adding a fully trained hidden layer to an ESN.

We conclude that the SpaRCe model considerably improved the performance and the convergence time of a reservoir on this biologically inspired benchmark task, with the relatively small overhead of $N$ additional parameters, one per reservoir node. 

\begin{center}
\begin{table}[h]
\begin{tabular}{ |p{0.4\linewidth}p{0.4\linewidth}|  }
\hline
 \multicolumn{2}{|c|}{Storage capacity Task} \\
 \hline
 \hline
 $ \sigma$  &   $0.3$  \\
 $\Delta_t$ & $0.1 s $  \\
 $T$ 		& $ 0.3s $  \\
 $\delta t$  &  $0.01s $ \\
 \hline
 $\alpha $    & $0.1$ \\
 $\rho$ &   $0.95$  \\
 $N$ & $1000$ \\
 $\gamma$ & $1$ \\
 $p_{ER}$ & $0.001$ \\
 \hline 
 $\eta_{W}$ & $2\times 10^{-3}$  \\
 $\eta_{\theta}$ & $2\times 10^{-4}$ \\
 minibatch size & $20$ \\
 \hline
\end{tabular}
\caption{\small{The table reports the parameters defining the task, the hyperparameters of the ESN and the training hyperparameters for the storage capacity task.}}
\end{table}
\label{Bio_table}
\end{center}

\subsection{Threshold learning increases performance} \label{task_ML}

In this section we faced three variants of classification tasks using the MNIST dataset. 
Each image is fed into the network sequentially either one column at a time or one pixel at a time to make the task temporally dependent. Thus, one written digit corresponds to a sequence of $28$ time steps of a $28$ dimensional input in the column by column paradigm, or to a sequence of $728$ time steps of a one dimensional input.

\subsubsection*{MNIST, column by column} \label{MNIST}
 
The application of ESNs on this specific task was previously analysed in \cite{schaetti2016echo}, in which the original dataset was preprocessed and augmented by resizing and deforming the original images. Without such a preprocessing, the ESN could not outperform a simple perceptron \cite{schaetti2016echo}.     
In contrast, in this work we use the original dataset, without any additional transformations. 
In this experiment, the pixels of each image are fed to the ESN sequentially column by column; thus, the input signal $\textbf{s}(t)$ corresponding to an  example image is a 28-dimensional sequence of temporal length 28. In this case, $\tilde{\textbf{V}}=\mathcal{C}\Big( \big\{\textbf{V}(t)\big\}_{\forall t}\Big)$, meaning that all the dynamic of the recurrent network across time is used to define the read-out of the system.
The cost function adopted is a sigmoidal cross entropy
\begin{equation}
E=-\Big[\sum_j \tilde{y}_j log\Big(\sigma(y_j)\Big)+(1-\tilde{y}_j)log\Big(1-\sigma(y_j)\Big)\Big]
\end{equation}
which is analysed in B.2 (Appendix). 
The optimizer used is Adam \cite{kingma2014adam}.
We first tested the SpaRCe model with various initial sparsity levels (Fig.\ \ref{Figure5}A, different colours). Regardless of the initial condition, the final performance was similar, as was the final level of sparsity (the size of dots in Fig.\ \ref{Figure5}A shows the percentage of active nodes in the network). The average error achieved by SpaRCe is $1.9\%$. The hyperparameter values are given in Table \ref{ML_table}.
The model reaches performance levels comparable to those achieved using two-layer or three-layer neural networks with backpropagation \cite{deng2012mnist}. We note that convolutional neural networks are the best performing networks for this task and for images classification problems in general. The best performance corresponds to an error rate of $0.21\%$ on the MNIST dataset and it is achieved through a pool of five convolutional neural networks \cite{deng2012mnist}.
However, the task faced here with SpaRCe is more challenging than the common approach used to train neural networks on the MNIST dataset, in which the whole image is fed into the network at once. The performance of convolutional neural networks and of multilayer perceptrons are reported in Table \ref{ML_table} as a benchmark for the results obtained with ESN with SpaRCe.
To demonstrate the importance of thresholds in the SpaRCe model, we compared the performance of  SpaRCe  to that of an ESN without thresholds trained online with the same optimizer, using a learning rate optimized through grid search. SpaRCe outperformed the ESN in both classification accuracy and convergence time (Fig.\ \ref{Figure5}B, `MNIST'). We also found that the normalisation mechanism introduced thanks to SpaRCe stabilises the learning process for relatively higher learning rates vs standard ESNs. 

As a second task, we applied the same model to the MNIST dataset where the pixels of each image are reordered through a permutation of the data. Each image is again fed one column at a time. Again, SpaRCe outperformed the ESN without thresholds (Fig.\ \ref{Figure5}B, `pMNIST').

\begin{center}
\begin{table}[h!]
\begin{tabular}{ |p{0.2\linewidth}p{0.2\linewidth}|p{0.2\linewidth}p{0.2\linewidth}|  }
\hline
\multicolumn{4}{|c|}{Hyperparameters} \\
\hline
\hline
 \multicolumn{2}{|c|}{MNIST/sMNIST} &  \multicolumn{2}{c|}{psMNIST} \\
 \hline
$\alpha $   &  $0.17$   & $\alpha_2 $   &  $0.017$    \\
$\rho$      &  $0.97$   & $\rho_2$      &  $0.99$   \\
$N$         &  $1000$   & $N_2$         &  $500$   \\
$\gamma$    &  $0.1$    & $\gamma_{21}$ &  $0.15$   \\
$p_{ER}$    &  $0.01$   & $p_{ER_2}$      &  $0.01$   \\

\cline{3-4}
            &           & $\alpha_1 $   &  $1$    \\
            &           & $\rho_1$      &  $1$   \\
            &           & $N_1$         &  $300$   \\
            &           & $\gamma_1$    &  $1$   \\
            &           & $p_{ER_1}$      &  $0.01$   \\
 \hline
$\eta_W $   &  $2\times 10^{-3}$   & $\eta_W $   &  $2 \times 10^{-3}$    \\
$\eta_{\theta}$      &  $2 \times 10^{-4}$   & $\eta_{\theta}$      &  $2 \times 10^{-4}$   \\
minibatch   &  $20$   & minibatch   &  $20$    \\
\hline
\hline
 \multicolumn{4}{|c|}{Results} \\
 \hline
 \multicolumn{2}{|c|}{MNIST} &  \multicolumn{2}{c|}{psMNIST} \\
  \hline
$ESN$ & $95.2$ & $ESN^{2}$ & $94.7$\\
$SpaRCe$ & $98.1$ & $SpaRCe^{2}$ & $95.4$ \\
$NN$ & $98.5^{*}$ & $GRU$ & $95.4^{**}$ \\
$Conv.$  & $98.3/99.8^{*}$ & $LSTM$ & $89.9^{**}$ \\   
\hline
\end{tabular}
\caption{\small{Table of the hyperparameters for the three benchmark tasks in Section \ref{task_ML}. The hyperparameters for the psMNIST are double since two ESNs are used for this task (as the symbol $^2$ over SpaRCe highlights). The suffix one corresponds to the first reservoir and the suffix two to the second reservoir. $\gamma_{21}$ indicates the input gain of the adjacency matrix from the first to the second ESN. The asterisks $^*$ $^{**}$ indicate that the results are taken from \cite{deng2012mnist} and \cite{chandar2019towards} respectively. $NN$, $Conv.$, $GRU$ and $LSTM$ stand for multilayer perceptrons, convolutional networks, Gated Recurrent Unit and Long Short Term Memory respectively.}}
\label{ML_table}
\end{table}
\end{center}

\subsubsection*{psMNIST}

We next analysed the performance of the SpaRCe model and of ESNs in general on the psMNIST (permuted sequential MNIST) task, which has became a standard benchmark for recurrent neural networks \cite{chandar2019towards}. In this case, each image is reordered through a fixed permutation and fed into the recurrent network one pixel at a time. The task is particularly challenging because temporal dependencies must be learned between widely-separated time steps of an image.
Since each sequence is a succession of $784$ pixels, we decided to sample the dynamic of the reservoir across time at constant steps multiples of 28, defining $\tilde{V}=\mathcal{C}\Big( \big\{ \textbf{V}(t) : 28|t \big\} \Big)$, where $28|t$ indicates that $t$ has to be divisible by $28$. This sampling procedure at constant temporal time steps could be suboptimal compared to an approach where the most informative time steps are selected. However, an analysis of optimal sampling procedures goes beyond the scope of this paper. 

Our first attempts to solve the task with a single ESN gave performance comparable to or worse than a standard perceptron model trained on the whole image, because the  ESN could not simultaneously discover long time dependencies and quickly adapt to new inputs. Indeed, in any network of randomly-connected low pass filters such as ours, it can be difficult to associate events that are distant in time, as it may be impossible to find a workable balance in the trade-off between keeping a memory of past events (which requires each node's activity to have a long time constant, i.e., slow decay) vs. allowing the network to evolve dynamically over time (which requires a short time constant, i.e., fast decay). 

To overcome this difficulty, we used an architecture composed of two reservoirs. The first has $300$ nodes with fast time constants, and the second has $1000$ nodes with slower time constants (the parameters for the two reservoirs are reported in Table \ref{ML_table}). The faster reservoir signals unidirectionally to the slower one, which is used for learning. This type of structure, where the timescales of the first network are faster than the second, was found to be optimal in previous works \cite{manneschi2020} . The sparsity level in the connectivity of the faster reservoir is important: it regulates a trade-off between too much connectivity (the relation between the input information to the second reservoir and the input signal becomes too complex) vs. not enough connectivity. The best performance arises when the network has the shortest path lengths between two nodes that could permit a sufficient amount of memory. 

\begin{figure*}[h!]
 \centering
    \makebox[\textwidth][c]{\includegraphics[width=0.95\textwidth]{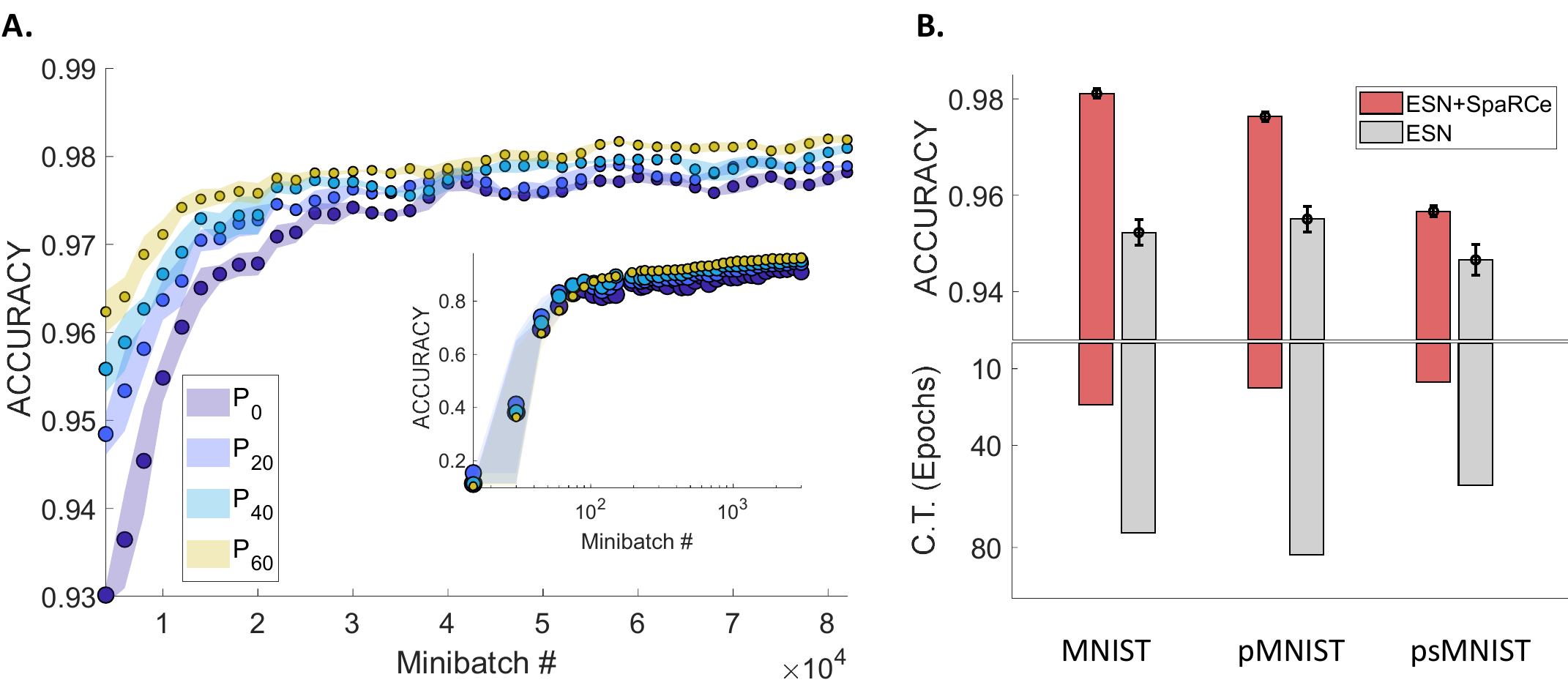}}
    \caption{ \small{The SpaRCe model shows comparable performance to a $2/3$ hidden layer neural network on the MNIST dataset and accuracy comparable to more complex RNNs trained with BPTT on the psMNIST task. On the contrary, the adoptation of the standard ESN read-out leads to lesser performance. \textbf{A}: The sizes of the dots reflect the percentage of active nodes (sparser network=smaller dots) in the network. Each minibatch corresponds to the presentation of $20$ training samples. The abscissa of the inset figure is scaled logarithmically. \textbf{B}: Performance of ESN with (red) and without (grey) threshold learning on the three tasks analysed, measured by accuracy and convergence time (C.T.). The network with the SpaRCe model outperforms the standard ESN read-out on all the benchmarks, but the contribution of the thresholds decreases as the task becomes more complex. This can be understood by considering that the increasing complexity of the tasks from left to right of the graph arises from a greater demand of the network's ability to understand long term dependencies. This aspect depends on the system dynamics and is not strongly related to threshold learning. Furthermore, the SpaRCe model converges about 5 times faster than an ESN without thresholds.  }}
  \label{Figure5}
\end{figure*}

Using this architecture, SpaRCe performs comparably to published methods despite using a much simpler training algorithm. As with MNIST and pMNIST, on psMNIST SpaRCe outperformed the threshold-less ESN in both accuracy and speed (Fig.\ \ref{Figure5}B). To compare SpaRCe with published models, we repeated the simulation with an ESN with $N=500$ nodes (thus $154\times10^{3}$ parameters), which gave an accuracy $>0.95$, comparable to the best, more complex recurrent neural networks that exploit backpropagation through time (BPTT), whose performances are 0.954 (GRU) and 0.899 (LSTM) \cite{chandar2019towards} (Table \ref{ML_table}) with a comparable number of parameters ($\approx167\times10^3$). While BPTT needs to unroll all the dependencies of the neural network activities across time, ESNs have the advantage to train a much simpler perceptron on top of the reservoir representation.   In particular, considering that the dynamic of a reservoir across datasets can be computed once only and then used to train a high dimensional perceptron, the computational cost of ESNs is reduced to the computational cost of RNNs.  

We emphasize that the procedure of concatenating previous temporal representations is not simply a shortcut, but it is necessary to increase the dimensionality of the representation in order to solve complex machine learning tasks through a reservoir computing approach. Indeed, the idea behind reservoir computing is to exploit the temporal dynamic of a system as a fixed and higher dimensional representation that allows it to separate the classes of a classification task through an hyperplane. This approach is antithetical to the learning process of a recurrent neural network with backpropagation through time, which trains the dynamics of the system and draws a nonlinear manifold to solve the classification task. Furthermore, the concatenating procedure does not guarantee any understanding of the long temporal dependencies among pixels that are necessary to effectively solve the problem. It is therefore of interest that ESNs using SpaRCe could perform comparably to state of the art recurrent networks, whose parameters are trained via a far more complex algorithm, backpropagation through time.


\subsection{SpaRCe alleviates catastrophic forgetting} \label{Catastrophic}

Catastrophic forgetting refers to the inability of standard neural networks to learn different tasks sequentially. If a neural network is trained on a specific dataset and then retrained to perform a novel task, it  
will probably forget what it has learned before. 
This unsolved problem \cite{kemker2018measuring} is critical for the future development of neural networks in general. 
Previous research formulated models that mitigate catastrophic forgetting, using a variety of techniques categorised by Kemker et al. \cite{kemker2018measuring}. These techniques include regularization  (impose a cost to changing the weights that contribute to previous tasks, as in Elastic Weight Consolidation, or EWC), rehearsal (re-playing previously learned data during subsequent training, as in GeppNet), and sparse coding (reducing the fraction of active nodes, as in the Fixed Expansion Layer model, or FEL, and Hard Attention to the Task, or HAT \cite{serra2018overcoming}). Sparse coding is also the approach we use here, with SpaRCe. However, in contrast to these techniques that exploit additional information, such as model awareness of the task identity, and the computation of ad-hoc quantities, such as the importance of specific parameters (EWC) or nodes (HAT) for a given task, we will demonstrate how the application of Eq.\ \ref{x} alone with a high starting sparsity level can alleviate catastrophic forgetting. Notably, the same  methodology applied in the previous simulations, to improve convergence time and performance of reservoir computing, can improve the ability of an ESN to learn different tasks sequentially. The difference between the application of SpaRCe in a single task and in a sequential tasks paradigm is the tuning of the hyperparameters of the proposed model, in particular the percentile value $n$ and the learning rates $\eta_W$ and $\eta_{\theta}$.
We will demonstrate how sparsity regulates a trade-off between initial learning speed vs.\ memory retention because high sparsity decreases overlaps among representations, which prevents new learning from disrupting old memories, whereas low sparsity means more nodes are active, allowing memories to be formed faster on new tasks. 

This trade-off between initial learning vs.\ preventing forgetting is studied in detail for two paradigms that are commonly used to measure catastrophic forgetting in neural networks:

\begin{itemize}
\item Sequential data permutations. A different permutation is applied to the considered dataset $N_{task}$ times. These new $N_{task}$  reshuffled datasets are then learnt sequentially by the system. In our simulation, $N_{task}=10$.
In such a scenario, the complexity of the different datasets is the same, and we trained SpaRCe for approximately two epochs for each task.  
\item Sequential classes. The first task is composed by the data corresponding to half of the possible classes, while the other classes are trained sequentially. Since in the dataset considered there will be $10$ classes, the number of tasks that the network has to learn sequentially is six. In this case, we trained the model for approximately one epoch for each task.
\end{itemize}

\begin{figure*}[h!]
 \centering
    \makebox[\textwidth][c]{\includegraphics[width=0.85\textwidth]{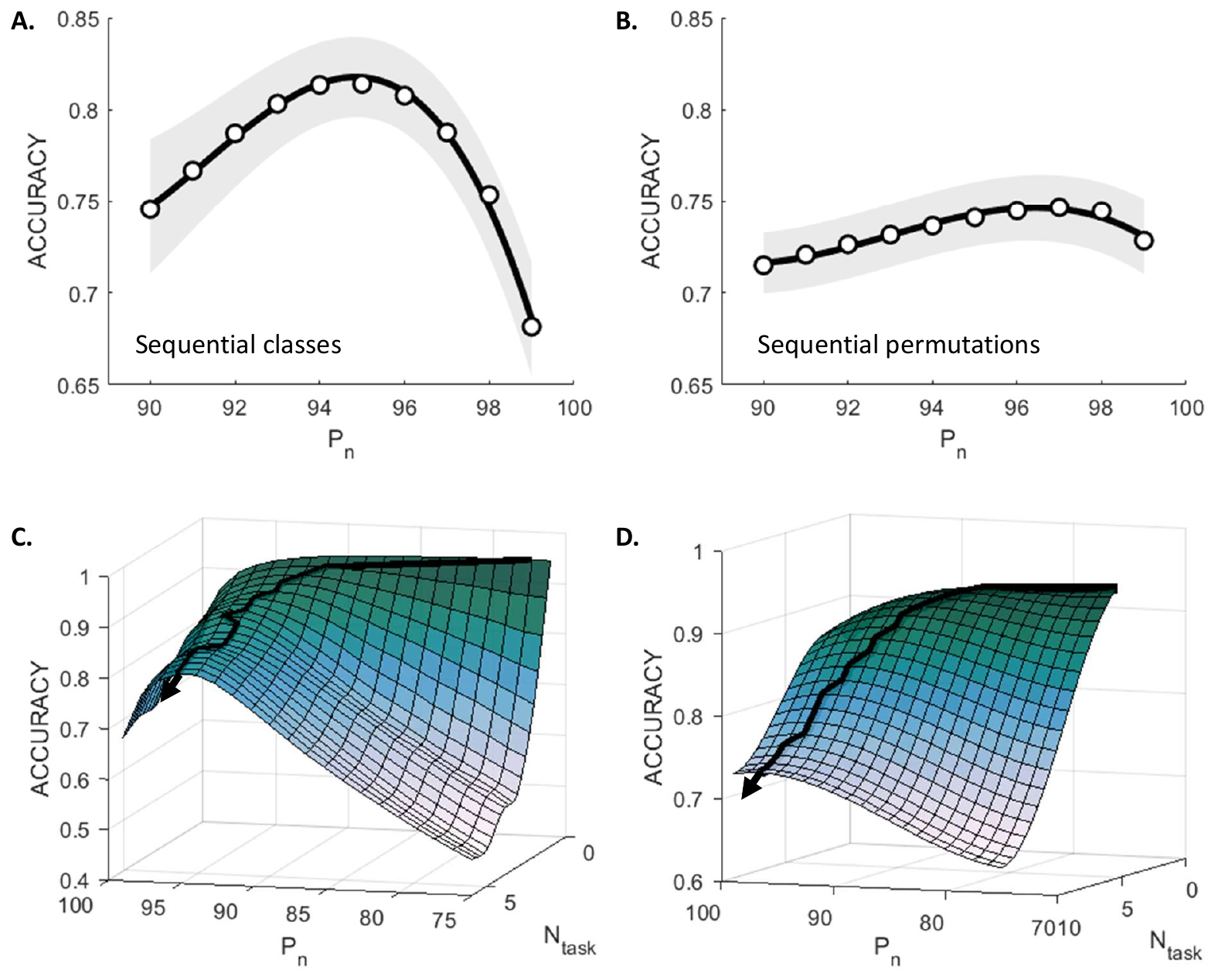}}
    \caption{\small{SpaRCe helps to prevent catastrophic forgetting on the two analysed  benchmarks. Different data points correspond to diverse repetitions of the experiment. \textbf{A}: Results on the MNIST dataset in the sequential classes paradigm (see text). \textbf{B}: Results on the permuted datasets paradigm.
\textbf{C}-\textbf{D}: Performance as function of the starting percentage of active nodes and the number of tasks that are learned in the catastrophic forgetting simulations (\textbf{C} refers to sequential classes and \textbf{D} for permuted datasets). The surface is a cubic interpolation of the accuracy as the number of datasets and the starting sparsity level vary. The path shows the best performing sparsity levels across various number of tasks; its movement from right to left demonstrates the necessity of adopting increasing level of sparsity as the number of datasets increases and the memory of previous tasks becomes more relevant.   }}
    \label{Catastrophic_ML}
\end{figure*}

The dataset used is MNIST, where the reservoir processed every image column by column as in Section \ref{MNIST}, adopting $\tilde{\textbf{V}}=\mathcal{C}\Big(\big\{\textbf{V}(t)\big\}_{\forall t}\Big)$ as before. Of course, the model is able to learn from the data corresponding to a task only once. We learn from the training dataset and compute the accuracy on the testing dataset by varying the initial starting sparsity levels through grid search of the value of $n$ in Eq.\ \ref{x} and as the number of tasks considered varies (Fig. \ref{Catastrophic_ML}C,D). These performance are used for comparison and demonstration only, and a validation dataset will be used to select the best hyperparamters' setting. 
In comparison to previous simulations, the learning rate for the thresholds is smaller (see Table \ref{Catastrophic_table}), because it was crucial to prevent a dramatic change of sparsity levels during learning, as such an abrupt change in the percentage of active nodes would alter stimulus representations and thereby affect previously learned tasks. In both tasks, lower sparsity levels allowed better initial learning on novel data (Accuracy across $N_{task}=1$, Fig.\ \ref{Catastrophic_ML}C,D), while higher sparsity levels alleviated forgetting of previous tasks during subsequent training. In other words, low sparsity allows good performance when the number of tasks is low, but the accuracy decreases quickly when $N_{task}$ increases. On the contrary, at optimal sparsity levels, accuracy remains high even as the number of tasks increases (highlighted path on the surface plot in Fig.\ \ref{Catastrophic_ML}C,D). These two conflicting trends  combined make the total accuracy across all datasets an inverted U when plotted against sparsity level (Fig.\ \ref{Catastrophic_ML}A,B). 

Moreover, we computed the following quantities to measure the alleviation of catastrophic forgetting in more details:
\begin{align}
\alpha_{Overall}=\dfrac{1}{N_{task}-1} \sum_{n=2}^{N_{task}} \dfrac{acc_n}{acc(1,1)}
\label{Cata_acc} \\
\alpha_{Memory}=\dfrac{1}{N_{task}}\sum_{n=1}^{N_{task}} [acc(n,N_{task})-acc(n,n)] \label{Cata_Memory}\\
\alpha_{New}=\dfrac{1}{N_{task}}\sum_{n=1}^{N_{task}} acc(n,n) \label{Cata_New}
\end{align}
where $acc_n$ is the accuracy computed on the datasets seen until task number $n$ (included), and $acc(1,1)$ is the accuracy of the first dataset immediately after its learning, which is considered as the ideal baseline. In general, we denote with $acc(n,m)$ the accuracy of the $n$-th dataset after the presentation of $m$ datasets.  The performance metric $\alpha_{Memory}$ measures model's ability to remember previous tasks, i.e., to alleviate catastrophic forgetting, while the metric $\alpha_{New}$ measures the model's ability to learn new tasks. In other terms, $\alpha_{Memory}$ is the average difference over tasks between the performance obtained after the whole training and immediately after the presentation of a specific dataset, while $\alpha_{New}$ is the average performance of a dataset after its presentation. These metrics and $\alpha_{Overall}$ are taken from \cite{kemker2018measuring}, where the performance of various models that alleviate catastrophic forgetting in multilayer perceptrons are compared. 

Finally, we repeated the simulation ten times for each of the two paradigms considered, averaged over different possible permutations (sequential data permutations) and for different ways of dividing the data into separate classes and corresponding tasks (sequential classes), selected the best performing algorithm based on the validation dataset and computed its performance $\alpha_{overall}$ on the test dataset.  The performance obtained is reported in Table \ref{Catastrophic_table} together with the results obtained by a standard ESN and published models that use a variety of strategies to prevent catastrophic forgetting in multilayer perceptrons. Of course, we do not expect to compete with newer Deep learning methods that exploit additional information and many more parameters. The results reported are obtained following, to the best of our knowledge, the same methodology as  \cite{kemker2018measuring}, and differ only in the necessity of an ESN to process the data temporally, rather than feeding the whole image to the network at once. In this aspect, the temporal processing of the data can make the task only more challenging for our model, since catastrophic forgetting appears to constitute an even more difficult problem to recurrent neural networks \cite{arora2019does}. We need to highlight, however, a drawback of our model: the tuning of an additional hyperparameter, that is the starting sparsity level $P_n$ regulated by the proposed normalisation mechanism. However, we note how all other methods that alleviate catastrophic forgetting in neural networks usually have additional hyperparameters (set through heuristics or grid search), as the learning rate of the penalty term in EWC.

If the results on multilayer perceptrons can be considered exclusively as a useful baseline, we notice from Table \ref{Catastrophic_table} the complete inability of an ESN to solve sequential learning. Indeed, the training of the read-out weights of an ESN in the standard reservoir computing paradigm causes a complete forgetting of previous tasks, resulting in an overall performance that corresponds to the learning of the last task processed by the model.

The success of SpaRCe on these catastrophic forgetting tasks arises from both the initial sparsity and from threshold learning. We analysed  the relative importance of the starting sparsity level vs.\ the online threshold adaptation introduced by the learning rule in the Supplementary Materials. 

\begin{center}
\begin{table}[t]
\begin{tabular}{ |p{0.2\linewidth}p{0.2\linewidth}|p{0.2\linewidth}p{0.2\linewidth}|  }
\hline
\multicolumn{4}{|c|}{Hyperparameters} \\
\hline
\hline
 \multicolumn{2}{|c|}{Sequential classes} &  \multicolumn{2}{c|}{Sequential permutations} \\
 \hline
$\alpha $   &  $0.17$   & $\alpha $   &  $0.17$    \\
$\rho$      &  $0.97$   & $\rho$      &  $0.97$   \\
$N$         &  $1000$   & $N$         &  $1000$   \\
$\gamma$    &  $0.1$    & $\gamma$ &  $0.1$   \\
$p_{ER}$    &  $0.01$   & $p_{ER}$      &  $0.01$   \\
 \hline
$\eta_W $   &  $5 \times 10^{-4}$   & $\eta_W $   &  $1 \times 10^{-3}$    \\
$\eta_{\theta}$      &  $5\times 10^{-5}$   & $\eta_{\theta}$      &  $1\times 10^{-5}$   \\
minibatch   &  $20$   & minibatch   &  $20$    \\
\hline
\hline
 \multicolumn{4}{|c|}{Results, $\alpha_{overall}$} \\
 \hline
 \multicolumn{2}{|c|}{Sequential classes} &  \multicolumn{2}{c|}{Sequential permutations} \\
  \hline
$ESN$ & $0.1$ & $ESN$ & $0.1$ \\
$SpaRCe$ & $0.870$ & $SpaRCe$ & $0.897$ \\
$EWC$ & $0.133^{+}$ & $EWC$ & $0.746^{+}$ \\
$FEL$  & $0.439^{+}$ & $FEL$ & $0.279^{+}$ \\
$GeppNet$  & $0.922^{+}$ & $GeppNet$ & $0.364^{+}$ \\
  
\hline

\end{tabular}
\caption{\small{Table of parameters used in the catastrophic forgetting tasks. We note that the learning rates used for the thresholds is smaller than in the previous simulations, since we needed to avoid changing the sparsity level quickly, in order to keep the optimised sparsity level near the starting value imposed ($P_n$). SpaRCe performs comparably to the best of the ad hoc models, which tend to perform well only on one of the two tasks analysed. The symbol $^+$ indicates that the results were taken from 
\cite{kemker2018measuring}.}}
\label{Catastrophic_table}
\end{table}
\end{center}

\section{Discussion}

It is customary in Machine Learning to introduce sparsity via regularisation: an ad-hoc penalty term is added to the network's error function to penalize the use of the weight parameters, leading to solutions with smaller (or sparser) weight values. In contrast, in neuroscience, sparsity is instead defined as the percentage of neurons that are active per stimulus, suggesting constraints not necessarily on the weights but rather directly on the neuronal activity. In this work, we take this latter approach. We learn a threshold per neuron via the minimisation of a standard error function, thereby associating sparsity directly to network performance. We  demonstrate theoretically that such a rule reduces the usage of reservoir neurons that have correlated activities and are connected to the output in the same way, that is with weights of the same sign. We also show theoretically, for the restricted case of positive weights (supplementary material), and in simulations, for the most general case (main text), that higher neuronal specialisation is a consequence of the standard error function applied on our network, where output neurons receive thresholded inputs from the reservoir. Due to threshold learning, neurons surviving the threshold will preferentially fire for one class vs another, which improves performance, as the search for a satisfactory solution takes place in a smaller search space. Indeed, we compare learning with and without thresholds on the same network, and find that thresholds improve both the speed and the accuracy of learning while the standard regularisation terms do not benefit the echo state network. Notably, threshold learning and weight learning in our setup is a two-way interaction: the threshold changes depend proportionally on the size of the weights. In practice, we have found that statistically an increase in neuronal specialisation follows large weight changes. 

We formulated the recurrent neural network by allowing one observable variable per neuron (the thresholded activity) and one hidden variable (the activity before the threshold). As such, learning the thresholds does not disrupt the dynamics of the neurons in the reservoir, which will evolve as in a standard echo state network. Instead, the output neurons will either receive input from the reservoir neurons or not, depending on the individual threshold of each of the reservoir neurons, which means that to learn the thresholds, we do not need to resort to backpropagation through time or other complex techniques. A biological interpretation of this structure could be that the recurrently connected neurons signal to each other based on subthreshold depolarization rather than action potentials. Such signalling could occur through dendro-dendritic synapses, which have been observed in the fly mushroom body \cite{zheng2018complete}, the structure that inspired the task in section \ref{task_bio}. 

Most interestingly, the learning rule we derive is structurally identical to the update rule for bias via backpropagation, since the learnable component of our formulation is effectively equivalent to adding a hidden layer containing as many neurons as the reservoir but with a very specific architecture: a one to one fixed connection to the neurons of the reservoir and a learnable threshold. The advantages of the proposed model (compared to adding a fixed weight, fully connected layer) arise from the specific architecture, the number of trainable parameters, the normalisation mechanism and the fact that weights and thresholds are learned in different time scales (learning rates). In our simulations, a hidden layer of 50 neurons was required to catch up with SpaRCe with approximately 47000 additional parameters (in comparison the 3000 used for SpaRCe), and it required longer training times (about twice slower).  
Another advantage of the threshold learning is that it helps stabilise the network if a large learning rate has been selected. In simulations we have observed that high learning rate values that lead to instabilities in the learning process for the non-threshold model lead to excellent performance for the threshold model: the thresholds act as a stabilisation mechanism, by quickly decreasing the activity of the network through a faster deactivation of neurons. As the mathematical analysis confirms, the threshold updates are proportional to the output weights, which suggests that thresholds move faster for larger weights. Furthermore, the simulations also indicate that the sparser the initial conditions the stronger the threshold changes, which again can be understood by the contradicting terms that a non-sparse network contributes to the threshold update. 

Notably, our model competes with feedforward and recurrent networks on standard benchmark problems (MNIST, sequential MNIST and permuted sequential MNIST), and is always best or close to the best alternative algorithm. This is a truly impressive result given the simplicity of our model and the relatively small overhead of the threshold process on the echo state network. While an echo state network is unlikely to compete in general with more complex networks that have a significantly higher number of learnable parameters, SpaRCe permits echo state networks to achieve performances that were not possible before, at least not without augmentation of the dataset before given to the network or combination with other, more complex machine learning algorithms.

Perhaps less obvious, the smart threshold initialisation is key to the success of the rule and its remarkably consistent performance, regardless of the exact threshold initialisation conditions. It is apparent from the mathematical formulae that the gradient rule for the threshold cannot activate silent neurons. Therefore, if the initialisation is entirely random, neurons with excessively high initial thresholds would never fire during the stimulus presentation.  Effectively, such neurons would be removed from the network for the whole duration of the simulation. To prevent this issue, the total input is first presented to the recurrent network, and we observe the operational activity range of each neuron. This allows us to set up a threshold within this regime, making sure that each neuron is active for a pre-decided percentage of time, across all stimulus presentations. In fact, one doesn't need to use the exact input of the network, but any signal(s) with the same statistics as the actual input. Similarly, it turns out that while some initial values may be better in terms of performance, in practice all that is needed is to give the same chance to all neurons to be active during the stimulus presentation, and the threshold learning will take care of the rest. 

Within the context of catastrophic forgetting, the last statement does not entirely hold. There is a clear advantage for our model if the sparsity level is initially set high, which we can also understand in the context of threshold updates being larger in magnitude for sparser networks. Our model competes with published models across two standard tests for catastrophic forgetting, unsurprisingly, given that we have shown that a consequence of our method is the increase of neuronal specialisation. In a sparse network different neurons are more likely to be used for different classes, and during the learning of new classes new neurons are likely to be recruited. In fact, sparsity without threshold learning significantly helps in the case of catastrophic forgetting, but threshold learning adds to the ability of better learning newer sets. In general, however, we do not expect to outperform complex new methods that exploit additional information: our model, following the principle of echo state networks and reservoir computing in general, uses inherent properties of the network (e.g. dynamics, sparsity) to boost performance in classification tasks and in catastrophic forgetting at a minimum computational cost. 

In summary, our work leads to a reinterpretation of the traditional role of thresholds in neural networks. We have shown that by disentangling the learning of the thresholds from the learning of the weights, and having a layer where learning takes place via threshold adaptation only, we were able to achieve sparse solutions and explain mathematically how this sparsity arises. To the best of our knowledge, this is the first time that this analysis and interpretation is provided for threshold learning, and we believe that this work might be applicable to network structures beyond echo state networks. 

Finally, reservoir computing is of increasing interest to the neuromorphic computing community, particularly to those who aim to use material dynamics for computation. For instance, in the spintronic community, magnetic devices are proposed as reservoir replacements, and more complex methods such as deep learning cannot be implemented in the material level. In the context of the echo state network, the reservoir serves only as a spatiotemporal kernel \cite{hermans2012recurrent}, i.e. it increases the dimensionality of the input signal in order to allow a linear model (a perceptron) to separate the classes. Therefore, it can be replaced by any highly non-linear but non-chaotic system that can transform its input to an appropriate higher dimensional space. Such proof of concept systems can be found for instance in \cite{markovic2019reservoir} \cite{romera2018vowel}. Our algorithm does not impose any modification to the reservoir itself, which allows its use even when the recurrent network is replaced by a physical material.

\section*{Acknowledgement}

We  thank  Paolo  Del  Giudice  and  Guido  Gigante  for  their input on the analysis of the timescales in the reservoir model, and Jelmer  Borst   for   suggesting   the   use   of   our   method on  catastrophic  forgetting.  EV  and  ACL  would  like  to  acknowledge  support  from  a  Google  Deepmind  Award.  EV was  partially  funded  by  the  Engineering  and  Physical  Sciences Research Council (Grant Nos. EP/S009647/1, EP/S030964/1 and EP/P006094/1). ACL was partically funded   by   the   European   Research   Council   (639489) and the Biotechnology and Biological Sciences Research Council (BB/S016031/1).

\clearpage
\section{Appendix} 

\subsection{Reservoir initialization} \label{sup1} 

The equation describing the dynamic of reservoir of leaky integrators is
\begin{equation}
\textbf{V}(t+1)=(1-\alpha)\textbf{V}(t)+\alpha f\big[\mathbf{W}_{in}\textbf{s}(t)+\rho \mathbf{W}\textbf{V}(t)\big]
\tag{A1} \label{V_sup}
\end{equation}
where $\mathbf{W}$ is a random connectivity matrix whose eigenvalues are uniformly distributed inside the unit circle of the imaginary plane, and $\rho<1$ is a constant.
The rescaling factor $\rho$ is called spectral radius and it is explicitly defined to control the maximum absolute value of the eigenvalues of the matrix $\rho \mathbf{W}$. The fact that the eigenvalues of the connectivity matrix $\rho \mathbf{W}$ are constrained inside the unit circle of the imaginary plane is a necessary condition for the Echo State property of the network.
Given the eigenvalues $\lambda_{\mathbf{W}}$ of $\mathbf{W}$, the eigenvalues $\lambda$ of the linearised dynamic system associated to Eq.\ \ref{V_sup} are
\begin{equation}
\lambda=(1-\alpha)+\alpha\rho \lambda_{\mathbf{W}} 
\tag{A2} \label{Eq.lambdas}
\end{equation}
and thus $\lambda_{\mathbf{W}}$ are compressed by a factor $\alpha$ and translated 
by a factor $1-\alpha$ in the imaginary plane. 
As a consequence, $\lambda$ follows the probability distribution
\begin{equation}
p(x,y) = \begin{cases} \dfrac{1}{\pi \alpha^2\rho^2}, & \mbox{if } 
\big[x-(1-\alpha)\big]^2+y^2 \leq \alpha^2\rho^2 \\ 
0, & \mbox{otherwise } \tag{A3} \end{cases}
\end{equation}
where $x=Re(\lambda)$ and $y=Im(\lambda)$ for simplicity of notation.
Since the real part of the eigenvalues is associated to the timescales $\tau$ of the dynamic system as $Re(\lambda)=\exp(-\dfrac{\delta t}{\tau})\approx 1-\dfrac{\delta t}{\tau}$, it is possible to compute the marginal distribution over $x$ of $p(x,y)$ for the real part, and then compute the distribution of timescales.
A simple strategy to choose $\alpha$ and $\rho$ by knowing the range of the timescales $[\tau_m,\tau_M]$ that the network should exhibit is to notice how the fastest (slowest) timescale $\tau_m$ ($\tau_M$) is given by the minimum (maximum) real eigenvalue of the dynamic system. 
Calling $\lambda_m=min\big\{Re(\lambda)\big\}$ and $\lambda_M=max\big\{Re(\lambda)\big\}$, and recalling we have

\begin{align*}
\lambda_m=1-\alpha + \alpha\rho~(-1) = 1- \alpha (1+\rho)=\\
=\exp(-\delta t/ \tau_m)  \approx 1- \dfrac{\delta t }{\tau_m} \rightarrow \\
\rightarrow \alpha (1+\rho) \approx\dfrac{\delta t }{\tau_m}
\end{align*}
and
\begin{align*}
\lambda_M=1-\alpha + \alpha \rho ~(+1) = 1-\alpha (1 -\rho)= \\
=\exp(-\frac{\delta t}{\tau_M}) \approx 1- \dfrac{\delta t }{\tau_M} \rightarrow \\
\rightarrow \alpha (1- \rho) \approx \dfrac{\delta t}{\tau_M}
\end{align*} 

Solving the system above, we end up with 

\begin{align*}
\alpha \approx\dfrac{\delta t }{2 \tau_m} + \dfrac{\delta t }{2 \tau_M}
\end{align*} 

and

\begin{align*}
\rho \approx\dfrac{\delta t }{2 \alpha \tau_m} - \dfrac{\delta t }{2 \alpha \tau_M}
\end{align*} 
that are relations between $\alpha$, $\rho$ and the minimum and  maximum timescales that the model can exhibit.
In this way, it is possible to choose the hyperparameters $\alpha$ and $\rho$ by selecting \textit{a priori} the more interpretable parameters $\tau_m$ and $\tau_M$. We want to emphasize that this procedure does not guarantee an optimal choice of the hyperparameters, but it can guide the search and it assures a good choice in terms of temporal memory of the reservoir.


\subsection{Thresholds initialization} \label{theta_initialisation}

Imposing a democratic initialisation where each node has the same probability to be active, the initial condition and sparsity level are defined by the choice of the starting percentile $P_n$. 
Here we defined two approaches to choose $P_n$:

\begin{itemize}
\item A simple grid search over $n$.  Here, $N_{P}$ reservoirs are trained in parallel for the first 10\% of time steps in the training instance, and the best performing reservoir is selected for the remainder of training. From our results, a small fraction of the training time is enough to choose the starting condition without any loss in the performance. 
\item Select the sparse representation that leads to the highest value of specialisation, a measure of the quality of the sparse representations that is defined below. 
\end{itemize}
The measure of \textit{specialisation} ($Sp$) reflects how a level of sparsity can facilitate the learning process in a classification task. The assumption behind the following formulation is that for a good sparse representation the ensembles of active nodes for different classes should overlap as little as possible.  
Let us consider two classes $j$ and $k$ and a neuron $i$. The node is specific if there is an asymmetry in the number of times it is active for one class with respect to the other. Generalizing this idea it is possible to build a measure $spec_{ijk}$ for a node $i$ defined as
\begin{equation}
spec_{ijk}=|\dfrac{N_{ij}}{M_j}-\dfrac{N_{ik}}{M_k}| \tag{A4}
\label{specialisation}
\end{equation} 
where $N_{ij}$ ($N_{ik}$) is the number of times the neuron $i$ was active after the presentation of a stimulus of class $j$ ($k$) and $M_{j}$ ($M_{k}$) is the total number of presentations of the stimuli belonging to class $j$ ($k$). 
Since the denominator of Eq.\ \ref{specialisation} contains the total number of presentations, $spec_{ijk}$ does not simply increase with the level of sparsity introduced. Let us focus on the particular case where $M_j\approx M_k$. A too high level of sparsity would lead the node to be almost silent, with a consequent poor specialisation value due to $N_{ij}$ and $N_{ik}$ being both close to zero. On the contrary, a too low sparsity level would lead the neuron to be excessively responsive, and $spec_{ijk}$ would be poor because $N_{ij} \approx N_{ik}$ even if $N_{ij}$ and $N_{ik}$ are both high. 

Given $spec_{ijk}$ it is possible to compute a measure of specialisation for each single neuron as
\begin{equation}
Sp_{i}=  \langle spec_{ijk} \rangle ^{(>0)}_{jk} \tag{A5}
\end{equation}  
where $\langle \cdotp \rangle^{(>0)}_{jk}$ is the average over positive elements for the indexes $jk$.  
It is possible to select the starting initial values of the thresholds as the n-th percentile of the distribution $\textbf{V}$ that leads to the highest specialisation measure.
Figure 3 (Main text) shows how the best performing sparse representation corresponds approximately to the maximum value of the average specialisation across neurons $Sp=\dfrac{1}{N}\sum_i^NSp_i$. 

\subsection{Gradients on thresholds \label{sup2}}

\begin{figure*}[h!]
 \centering
    \makebox[\textwidth][c]{\includegraphics[width=0.8\textwidth]{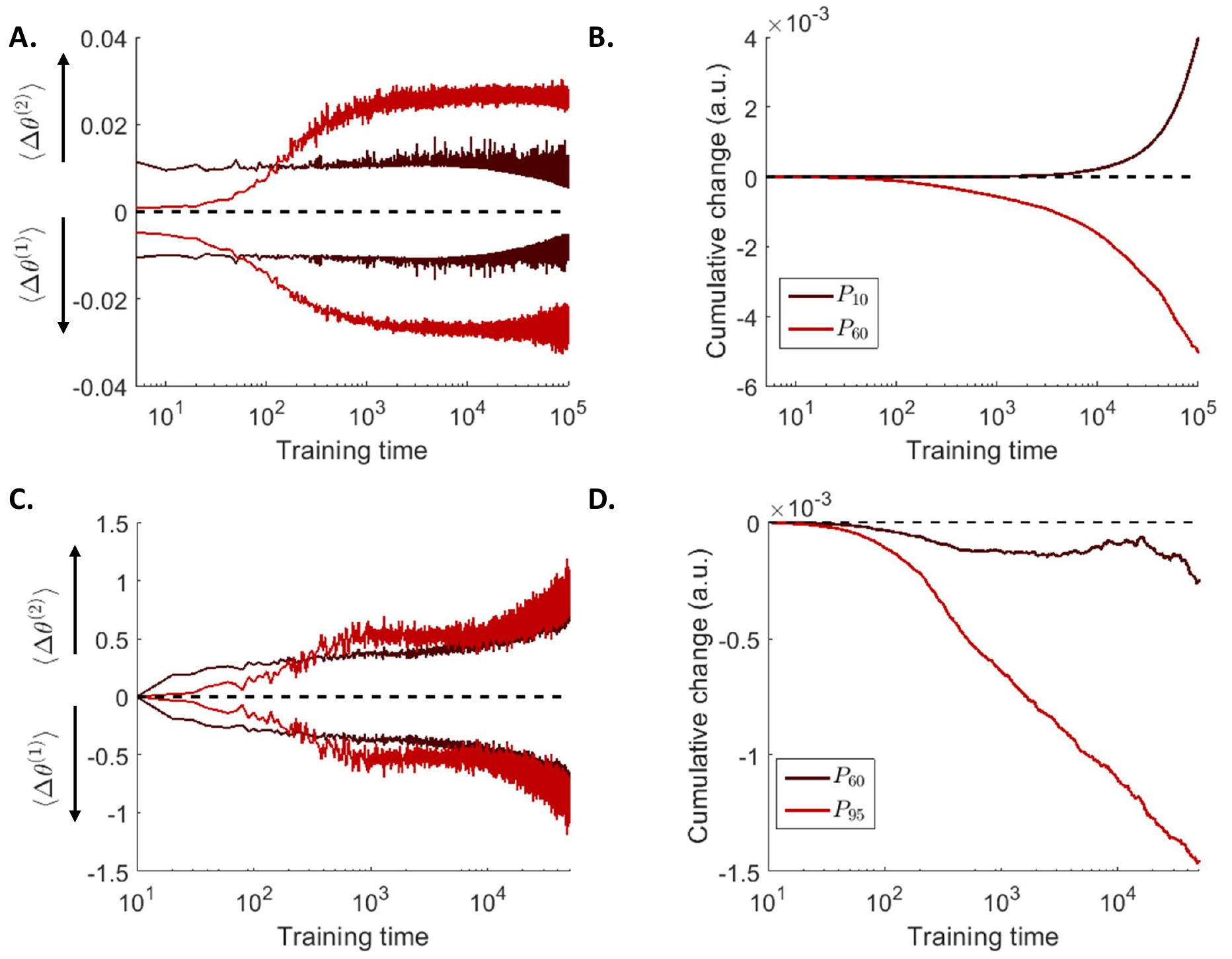}}
    \caption{\small{$\Delta \theta^{(1)}$ and $\Delta \theta^{(2)}$ drive the learning process of the thresholds. \textbf{A-C:}  Values of $\Delta \theta^{(1)}$ and $\Delta \theta^{(2)}$ averaged across the population of nodes for two different starting sparsity levels. These two factors have interpretable meaning (see Main text). \textbf{B-D:} Average of the cumulative threshold change. If the starting condition is suboptimal and low (high), such an average will be positive (negative) and consequently increasing (decreasing) the level of sparsity. Panels \textbf{A} and \textbf{B} refer to the storage capacity task of Section 3.1, where the cost function is a mean square error, while panels \textbf{C} and \textbf{D} refer to the classification of the MNIST dataset in the column by column paradigm of Section 3.2, where a sigmoid cross-entropy function was adopted. In the latter case, the two factors are defined in Eq. \ref{theta_+cross} and \ref{theta_-cross}.       }}
    
  \label{Forces_App}
\end{figure*}

\begin{figure*}[h!]
 \centering
    \makebox[\textwidth][c]{\includegraphics[width=1\textwidth]{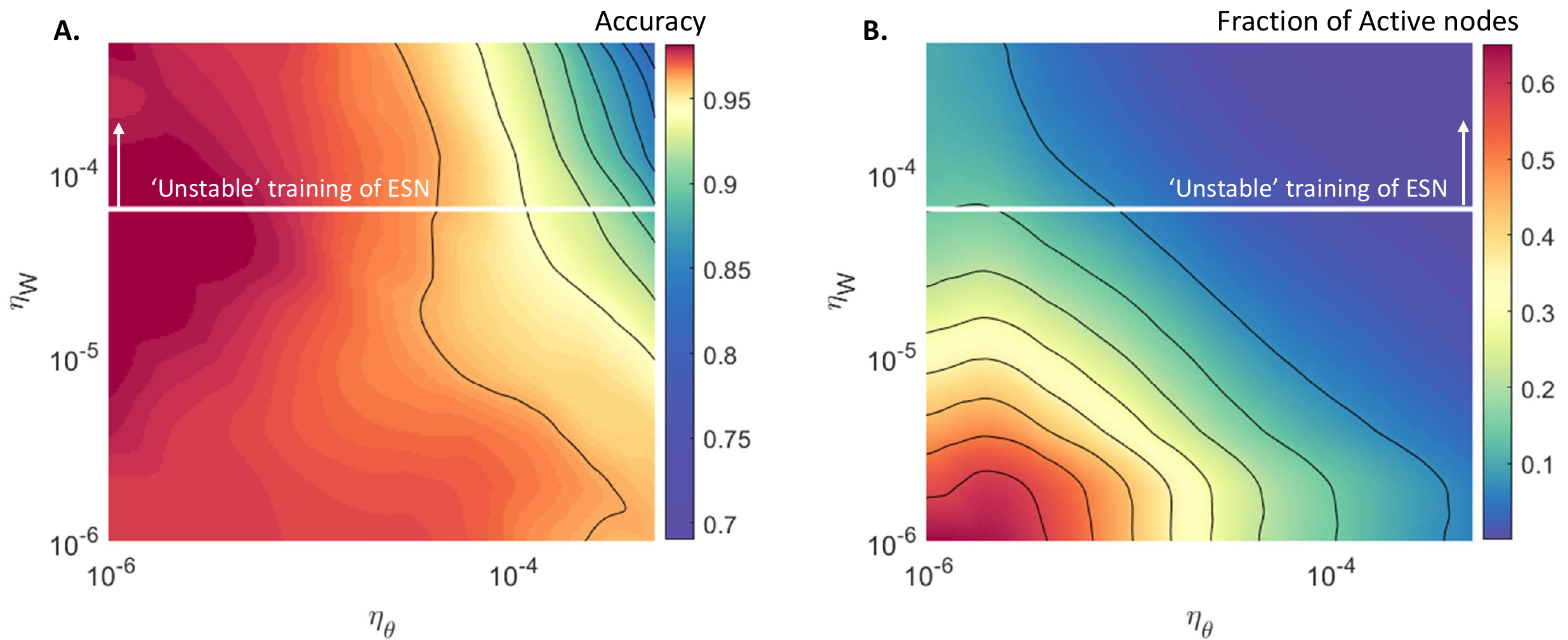}}
    \caption{\small{The proposed model acts as a stabilisation mechanism that permits the utilisations of higher learning rates in comparison to the values adopted in the standard ESN read-out. The result are obtained on the MNIST task of Section 3.2 in the column by column paradigm. For all the results shown, the model is initialised from a starting sparsity level $P_n=50$. We plot the accuracy (\textbf{A}) and the sparsity level (\textbf{B}) as the learning rates $\eta_{W}$ and $\eta_{\theta}$ vary.  \textbf{A}: The best performance of the algorithm corresponds to a region where the learning rate on the thresholds is at least $10^{-1}$ times lower than the learning rate on the output weights. This result is expected, considering that the learning rule on the thresholds depend explicitly on the output weights (Eq.\ \ref{theta_gradient}), and thus thresholds can be accurately learnt  when the output weights carry information on the classification process, i.e. weights are learnt faster than  thresholds. \textbf{B}: The percentage of active nodes decreases as the learning rates $\eta_{W}$ and $\eta_{\theta}$ increase. This can be understood considering that, when the learning rates are high, the model avoids abrupt changes on the output $\textbf{y}$ by decreasing the activities of the representation, i.e. increasing the sparsity level of the network. The parameter space above the white horizontal line  corresponds to a region where learning with the standard ESN read-out is unstable because of the too high value of $\eta_{W}$ adopted. In that region, training is characterised by an undesirable increase of the cost function across learning for the ESN but not for SpaRCe. }}
    
  \label{Rates}
\end{figure*}

The training procedure minimizes a measure of the distance $E(t)$ between the output $\textbf{y}=\mathbf{W}^{o}\textbf{x}$ of the neural network and the desired value $\mathbf{\tilde{y}}$. Theoretically, 
\begin{equation}
E=dist\Big(\mathbf{\tilde{y}},\textbf{y}\Big) \tag{B1}
\label{E_theory}
\end{equation} 
We will now apply a gradient based optimization on an example cost function, and show how the resulting learning rule for the thresholds can be interpreted.

\begin{flushleft}
\textbf{\textit{Gradient on $\bm{\theta}$, Mean Square Error (MSE)}}
\end{flushleft} 
Let us consider the mean square cost function, given by
\begin{align}
E=dist\Big(\mathbf{\tilde{y}},\textbf{y}\Big)=\nonumber\\
=\sum_j\Big[\tilde{y}_j-y_j\Big]^2=\nonumber\\
=\sum_j \Big[\tilde{y}_j-\sum_i W^{o}_{ji}x_i\Big]^2=\nonumber\\
=\sum_j \Big[\tilde{y}_j-\sum_i W^{o}_{ji}sign(\tilde{V}_i)relu\Big(|\tilde{V}_i|-P_n(|\tilde{\textbf{V}}_i|)-\tilde{\theta}_i\Big)\Big]^2 \tag{B2}
\label{E}
\end{align}
where we have used a read-out of Eq. 2 (Main text) to define the output of the neural network.
A gradient based approach that minimizes $E$ leads to the following learning rule on the output weights
\begin{align*}
\Delta W^{o}_{lk}=-\eta_W\frac{\partial E}{\partial W_{lk}}=\\
=\eta_W\Big[\tilde{y}_l-y_l\Big]x_k
\end{align*} 
and to the following learning rule for the thresholds
\begin{align}
\nonumber
\Delta \theta_k =-\eta_{\theta} \dfrac{\partial E}{\partial \theta_k}=-\eta_{\theta} \dfrac{\partial E}{\partial \tilde{\theta}_k}= \\\nonumber
= \eta_{\theta} \sum_{j=1}^{N_{class}} \big[ \tilde{y}_j-y_j \big]  \dfrac{\partial }{\partial \tilde{\theta}_k} \Big\{\sum_i W^{o}_{ji} x_i\Big\}=\\\nonumber
=-\eta_{\theta}\sum_{j=1}^{N_{class}} \big[ \tilde{y}_j-y_j \big] W^{o}_{jk}sign(\tilde{V}_k)H\Big(|\tilde{V}_k|+\\
-P_n(|\tilde{\textbf{V}}_k|)-\tilde{\theta}_k\Big)= \nonumber\\
= \eta_{\theta}\sum_{j=1}^{N_{class}} \big[ \tilde{y}_j-y_j \big] W^{o}_{jk}sign(x_k) = \nonumber\\
=-\eta_{\theta}\sum_{j=1}^{N_{class}} \tilde{y}_j W^{o}_{jk}sign\big(x_k\big) +\nonumber\\
+\eta_{\theta}\sum_{j=1}^{N_{class}}  y_j W^{o}_{jk}sign\big(x_k\big) \tag{B3}
\label{theta_gradient}
\end{align}

By taking into account the specific case of a classification task where $\tilde{y}_j$ is positive for $j$ that corresponds to the desired class and zero otherwise, it is possible to manipulate Eq.\ \ref{theta_gradient} and to separate it in two terms to uncover the meaning of the learning on the thresholds.  

\begin{align}
\Delta \theta_k= -\eta_{\theta} \beta W^{o}_{\tilde{j}k}sign\big(x_k\big)+\nonumber\\ 
+\eta_{\theta}\sum_{j=1}^{N_{class}}  y_j W^{o}_{jk}sign\big(x_k\big)  \nonumber\\
=-\eta_{\theta}\beta W^{o}_{\tilde{j}k}sign\big(x_k\big)+\nonumber\\
+\eta_{\theta}\sum_{j=1}^{N_{class}}\sum_{l=1}^{N}W^{o}_{jl}W^{o}_{jk} x_l sign\big(x_k\big)
\tag{B4}
\label{theta_gradient_mean}
\end{align}
where $\tilde{j}$ indicates the correct class for the considered input, and $\beta$ is the positive quantity equal to the correct desired output value $\tilde{y}_{\tilde{j}}$. Eq.\ \ref{theta_gradient_mean} contains two clearly interpretable factors:
\begin{align}
\Delta \theta^{(1)}=\sum_{j=1}^{N_{class}}\sum_{l=1}^{N}W^{o}_{jl}W^{o}_{jk}x_lsign\big(x_k\big) \label{theta_+} \tag{B5} \\ 
\Delta \theta^{(2)}=-\beta W^{o}_{\tilde{j}k}sign\big(x_k\big) \label{theta_-} \tag{B6}
\end{align}

\begin{flushleft}
\textbf{\textit{Gradient on $\bm{\theta}$, cross entropy}}
\end{flushleft}

The error function has the form
\begin{equation}
E=-\Big[\sum_j \tilde{y}_j log\Big(\sigma(y_j)\Big) +(1-\tilde{y}_j)log\Big(1-\sigma(y_j)\Big) \Big] \nonumber
\end{equation}

In this case, the learning rule for the thresholds is
\begin{align}
\Delta \theta_{k}=-\eta_{\theta} \sum_j \tilde{y}_j\big(1-\sigma(y_j)\big)W_{jk}sign(x_k)+ \nonumber \\
-\eta_{\theta} \sum_j \big(1-\tilde{y}_j\big)\sigma(y_j)W_{jk}sign(x_k)= \nonumber \\
=-\eta_{\theta}\sum_j \tilde{y}_j W_{jk} sign(x_k)+ \nonumber \\
+\eta_{\theta}\sum_j y_j \sigma(y_j) W_{jk} sign(x_k) \tag{B7}\label{two_terms_sig}
\end{align}
The two terms in Eq.\ \ref{two_terms_sig} have comparable meaning to $\Delta \theta^{(2)}$ and $\Delta \theta^{(1)}$ of Eq.\ \ref{theta_+} and \ref{theta_-} computed for the mean square error, To demonstrate this, we can consider the case of a classification task where $y^{true}_j=1$ for the correct class and zero otherwise. Furthermore, considering that the neural network output is not in the saturating regime of the sigmoid function when the majority of the learning happens, we can use the dominant first term of the Taylor series of the sigmoid and approximate the second term of Eq.\ \ref{two_terms_sig}
\begin{align}
\Delta \theta_{k}=-\eta_{\theta}W_{\tilde{j}k} sign(x_k)+\nonumber\\
+\eta_{\theta} \sum_j\sigma(y_j)W_{jk}sign(x_k)=\nonumber \\
=-\eta_{\theta}W_{\tilde{j}k} sign(x_k)+\nonumber\\
+\eta_{\theta}\sum_j\big[\dfrac{1}{2}+\dfrac{1}{4}y_{j}+...\big]W_{jk}sign(x_k)=\nonumber \\
=-\dfrac{1}{2}\eta_{\theta}W_{\tilde{j}k}sign(x_k)+\dfrac{1}{2}\eta_{\theta}\sum_{j\neq\tilde{j}}W_{jk}sign(x_k)\nonumber\\
+\eta_{\theta}\sum_j \big[\dfrac{1}{4}y_j+...\big] W_{jk} sign(x_k) \tag{B8}
\end{align}
where we can define 
\begin{align}
\Delta \theta^{(2)}=-\dfrac{1}{2}\eta_{\theta}W_{\tilde{j}k}sign(x_k)+\dfrac{1}{2}\eta_{\theta}\sum_{j\neq\tilde{j}}W_{jk}sign(x_k)  \label{theta_-cross} \tag{B9} \\
\Delta \theta^{(1)}=\eta_{\theta}\sum_j \big[\dfrac{1}{4}y_j+...\big] W_{jk} sign(x_k) \label{theta_+cross}\tag{B10} 
\end{align}
Eq. \ref{theta_-cross} has an analogous meaning to Eq.\ \ref{theta_-} with the additional term  $\sum_{j\neq\tilde{j}}W_{jk}sign(x_k)$ that increases (decreases) thresholds of nodes that are helping (impeding) the wrong classification process. Considering only the linear term of the Taylor expansion of the sigmoid function, Eq. \ref{theta_+cross} has the exact same form as Eq. \ref{theta_+}. To demonstrate its role as balancing term that deactivates nodes that are helping in a similar way the classification process (their contribution has the same sign of $y_j$), we estimated Eq. \ref{theta_+cross} by subtracting $\Delta \theta^{(2)}$ to the value of the gradient. The result of this procedure is shown in Panel C of Fig. \ref{Forces_App} (Supplementary).
\newline \newline

Finally, the sparsity level reached after optimisation depends on the values of the learning rates $\eta_{W}$ and $\eta_{\theta}$ used. Indeed, threshold learning can act as a balancing force that, in the case of high learning rates, decreases the percentage of active nodes in the network  and stabilizes the training phase. In this regard, Fig. \ref{Rates} (Supplementary) shows the accuracy and the corresponding sparsity level after optimisation as $\eta_{W}$ and $\eta_{\theta}$ vary for the MNIST classification task faced in Section 3.2 where the input is given to the network column by column (as a 28-th dimensional sequence of 28 time steps).

\subsection{Initialisation and threshold adaptation in sequential learning}

\begin{figure*}[h!]
 \centering
    \makebox[\textwidth][c]{\includegraphics[width=1\textwidth]{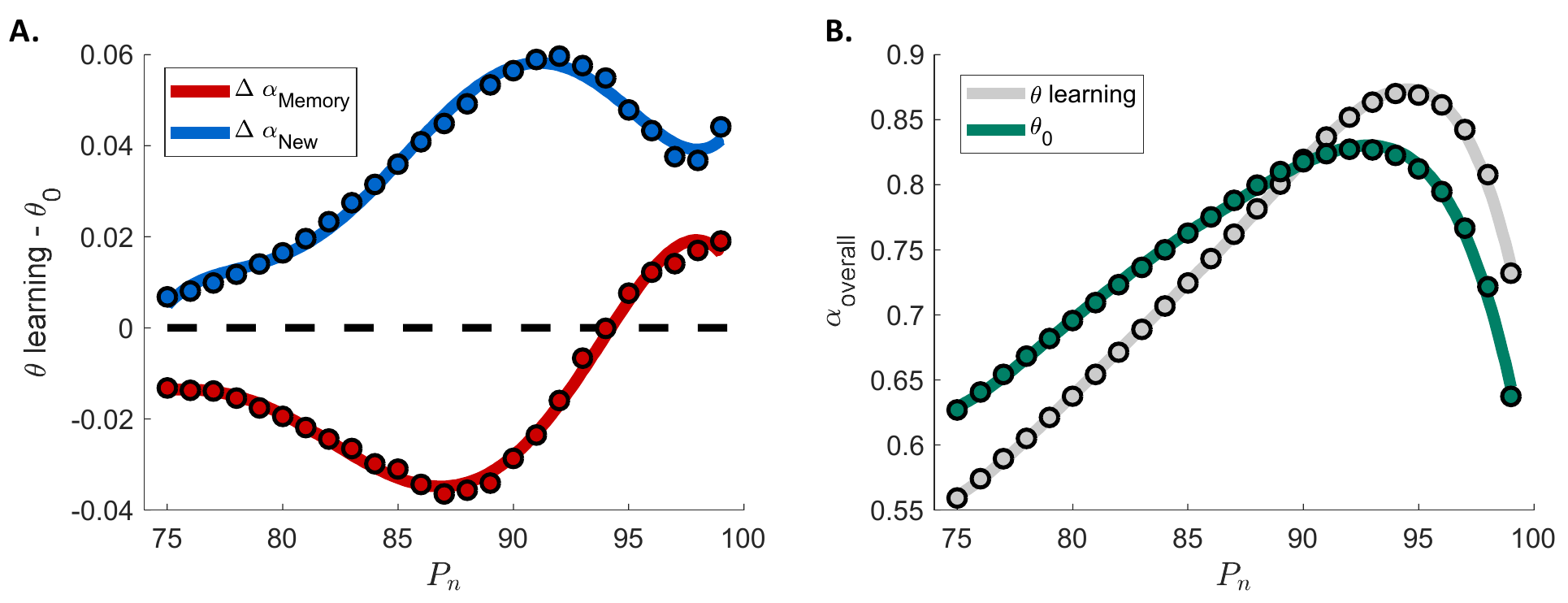}}
    \caption{\small{ The initialisation procedure is the key element to alleviate catastrophic forgetting. \textbf{A}: Difference between the full SpaRCe model and a network where the thresholds are exclusively initialised for two metrics (Eq. \ref{Delta_alpha_memory} and \ref{Delta_alpha_new}), which measure the ability of the network to remember (red) and to learn novel tasks (blue) respectively. \textbf{B}: Overall accuracy with and without thresholds adaptation, which leads to an increase in the maximum performance of approximately five percent. The high accuracy reached exclusively with the initialisation demonstrates its importance to alleviate catastrophic forgetting.    }}
    \label{Catastrophic_overlap}
\end{figure*}

We used Eq. 10, 11 and 12 introduced in the Main Text to calculate the differences 
\begin{align*}
\Delta \alpha_{Memory}= \alpha_{Memory}-\alpha_{Memory,\bm{\theta}_0} \tag{B11} \label{Delta_alpha_memory}\\
\Delta \alpha_{New}= \alpha_{New}-\alpha_{New,\bm{\theta}_0} \tag{B12} \label{Delta_alpha_new}
\end{align*}
between a model where thresholds are initialised and then learned and a model where the thresholds are initialised only (as marked by the subscript $\bm{\theta}_0$). Up to high sparsity levels, threshold learning reduces the model's ability to remember previous tasks by changing and compromising the parameters optimised on past datasets (seen as $\Delta \alpha_{Memory} < 0$ in Fig.\ \ref{Catastrophic_overlap}A, Supplementary). However, at initial sparsity above $P_n=94$, the initial memory capacity of the network is reduced by the small number of active nodes; threshold learning helps to recruit inactive neurons and to increase the amount of resources available. This leads to a consequent increase of $\alpha_{Memory}$ for high sparsity levels. At the same time, threshold optimisation facilitates the learning of new datasets by adapting the parameters to the new incoming stimuli ($\Delta \alpha_{New} > 0$, Fig.\ \ref{Catastrophic_overlap}A, Supplementary, blue trend). Altogether, the trade-off between memory and adaptability on novel data results in the trend seen in Fig.\ \ref{Catastrophic_overlap}B (Supplementary): at lower initial sparsity levels, threshold learning worsens overall performance (because it reduces $\alpha_{Memory}$), while at higher initial sparsity levels, threshold learning improves overall performance (because high sparsity prevents threshold learning from worsening $\alpha_{Memory}$ while still allowing it to improve $\alpha_{New}$). In summary, threshold learning modulates the starting condition and increases the maximum performance ($\alpha_{overall}$) by about five percent.

\subsection{Procedure for building sequences \label{sup3}}

\begin{figure*}[h!]
 \centering
    \makebox[\textwidth][c]{\includegraphics[width=0.98\textwidth]{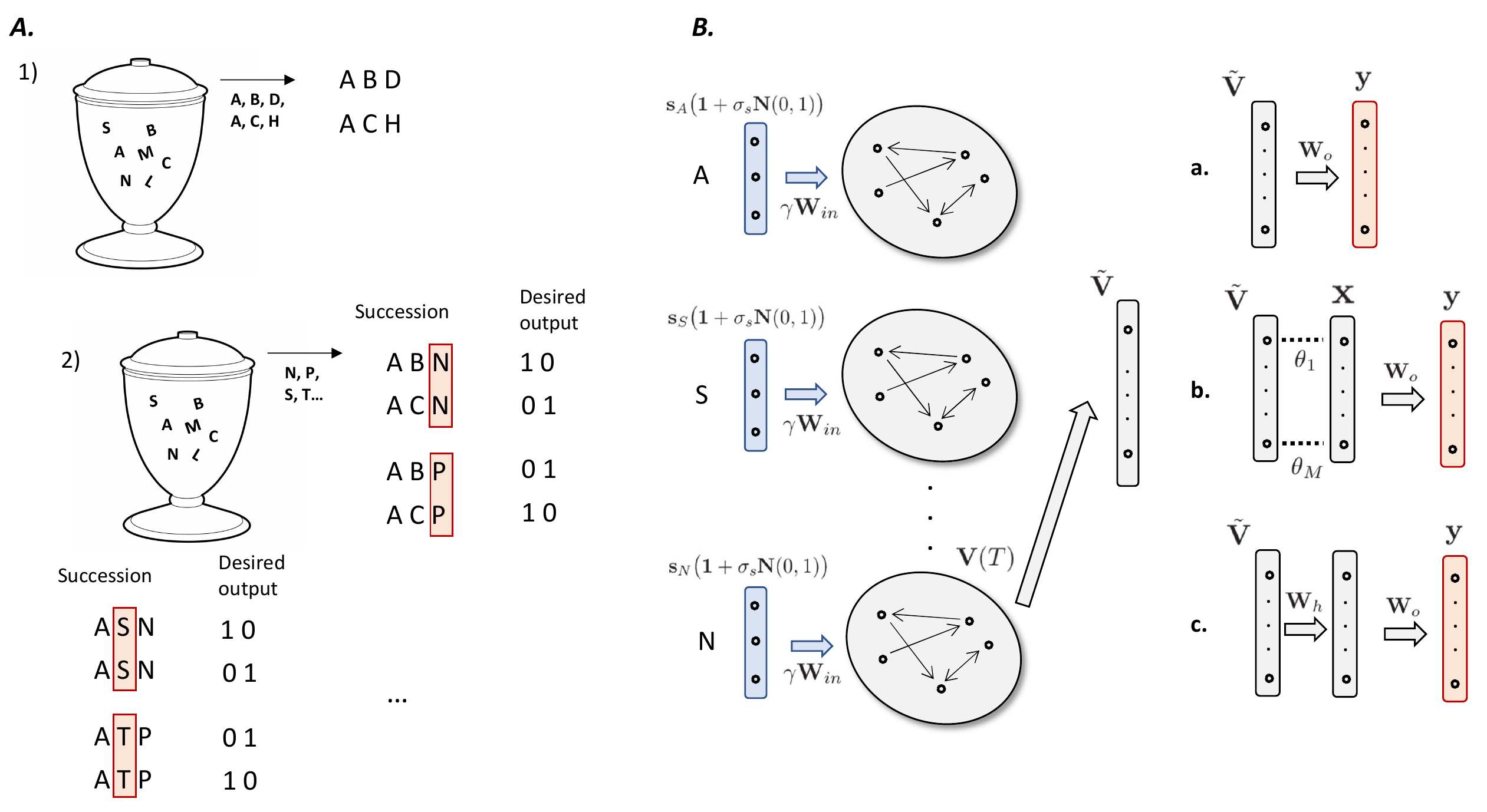}}
    \caption{\small{Task diagrams. \textbf{A:} Schematic of the procedure to define sequences. We notice how similarities among sequences are associated to different desired outputs, making the task a test of the storage capacity of the network. \textbf{B:} Schematic of the task on the biological data. Input example, succession of three stimuli (A,S,N) of time duration $\Delta t=0.1 {s}$ each. Multiplicative white noise $\sigma_s \mathcal{\textbf{N}}(0,1)$ is added throughout all the sequence, with different realisations of $\mathcal{\textbf{N}}(0,1)$ (gaussian with zero mean and unitary variance) for each temporal step. Only the last activities of the nodes in the reservoir is used for the read-out, which is divided into three models (\textbf{a}, standard ESN, \textbf{b}, SpaRCe, \textbf{c}, additional hidden layer) of increasing complexity and whose performance are tested and reported in the main text. }}
    
  \label{Sequence}
\end{figure*}

Given an ensemble of elements $\mathcal{E}=\big\{A,B,C,...\big\}$, we formulated a systematic procedure to build sequences of $N_t$ (in the case analysed, $N_t=3$) elements from $\mathcal{E}$. Our goal is to test the storage capacity of the model to learn associations between sequences and desired outputs. To achieve this, we prevented correlations between similar sequences from helping classification, by placing similar sequences in different classes. As described in Fig. \ref{Sequence}A (Supplementary), the procedure is based on the repetition of the following two steps:
\begin{itemize}
\item[1.] Given a number of output classes $N_{class}$, we picked  $N_tN_{class}$ random elements with repetitions from $\mathcal{E}$ and composed the successions as in Fig.~\ref{Sequence} (Supplementary), where the simple case of two output classes is considered. \newline
\item[2.] We picked other $N_tN_{class}$ random elements with no repetitions. Then, we changed the last temporal elements of the sequences generated before with $N_{class}$ of the new picked stimuli, associating the new sequences to different desired output values as shown in the figure. Finally, we proceed in this way for all the previous temporal elements of the considered $N_{class}$ successions from which we started (point 1). 
\end{itemize}

We notice how the similarities between sequences can not be used to infer the right classification, since correlations among elements are associated to different output classes. Each element is associated to a multidimensional signal that, with addition of multiplicative white noise, is presented to the network (Fig. \ref{Sequence} B, Supplementary).

\subsection{Necessity of sparse representation for obtaining an optimal solution with positive output weights}

The following section contains a formal proof of how sparsity is necessary to find an optimal solution of a classification problem when the output weights are constrained to be positive.
The cost function $E$ in a classification task and for the entire dataset can be written as:
\begin{equation}
E=\sum_{k}\sum_{\mu \varepsilon  \mathcal{C}_{k}}\sum_{j}^{N_{class}}\big[\tilde{y}_j^{\mu_{k}}-y_j^{\mu_{k}}\big]^2 \tag{D1}
\end{equation}
where $k$ indicates the correct class and the expression $\mu$ $\varepsilon$ $\mathcal{C}_{k}$ can be read as all datapoints $\mu$ that belongs to the cluster $\mathcal{C}_{k}$ from which we desire output $k$. We assume that the target is 1 for the correct class and 0 otherwise. An ideal solution of this problem with an online learning learning algorithm can be defined as:
\begin{equation}
\forall \, \varepsilon >0, \; \exists \, n \; | \; E_n \leq \varepsilon \tag{D2}
\label{Ideal_sol} 
\end{equation}
where $n$ is the minibatch (or batch) number and it is a measure of the training time.
Thus, Eq.\ \ref{Ideal_sol} means that the cost function can be made as small as desired by going through enough training instances. The fact that the sum of the quadratic terms of the cost function has to be less than or equal to a desired $\epsilon$ value means that 
\begin{align*}
[\tilde{y}_j^{\mu_{k}} - y_{j}^{\mu_{k}}]^2 \leq \varepsilon, \; \forall \mu, \; \forall k \\
\rightarrow -\sqrt{\varepsilon}\leq \tilde{y}_j^{\mu_{k}} - y_{j}^{\mu_{k}} \leq \sqrt{\varepsilon}, \; \forall \mu, \; \forall k 
\end{align*}
Thus, for each datapoint $\mu_k$ and each output class $j$  

\begin{equation*}
\begin{cases} 1-\sqrt{\varepsilon}\leq y_{j}^{\mu_{k}} \leq 1+\sqrt{\epsilon}, & \mbox{if } j=k \\  -\sqrt{\varepsilon}\leq y_{j}^{\mu_{k}} \leq \sqrt{\epsilon} & \mbox{if } j\neq k \end{cases}
\end{equation*}
which, by considering separately all different classes, becomes
\begin{equation}
\begin{cases} 1-\sqrt{\varepsilon}\leq y_{1}^{\mu_1} \leq 1+\sqrt{\epsilon},  \\  
-\sqrt{\varepsilon}\leq y_{1}^{\mu_k} \leq \sqrt{\epsilon} & \mbox{if } k\neq 1  \\
1-\sqrt{\varepsilon}\leq y_{2}^{\mu_2} \leq 1+\sqrt{\epsilon},  \\  
-\sqrt{\varepsilon}\leq y_{2}^{\mu_k} \leq \sqrt{\epsilon} & \mbox{if } k\neq 2\\
\hphantom{xxxxxxxxxx}\vdots \\
1-\sqrt{\varepsilon}\leq y_{C}^{\mu_{C}} \leq 1+\sqrt{\epsilon},  \\  
-\sqrt{\varepsilon}\leq y_{C}^{\mu_{k}} \leq \sqrt{\epsilon} & \mbox{if } k\neq C \tag{D3}\\
\end{cases}
\label{Case_}  
\end{equation}
Trivially, the difficulty of the classification task lies in the requirement that the activity of the same output node has to be close to one for some datapoints and close to zero for others; this condition is highlighted explicitly by the above set of inequalities, which can be rewritten 
\begin{equation*}
\begin{cases} 1-\sqrt{\varepsilon}\leq \sum_{l} W^{out}_{1l} relu(V_l^{\mu_1}-\theta_l) \leq 1+\sqrt{\epsilon},  \\  
-\sqrt{\varepsilon}\leq \sum_{l} W^{out}_{1l} relu(V_l^{\mu_k}-\theta_l) \leq \sqrt{\epsilon}, \;  \mbox{if } k\neq1  \\
\hphantom{xxxxxxxxxx}\vdots \\
\end{cases}
\label{Case_2}  
\end{equation*}
Since we required positive output weights we find 
\begin{align}
-\sqrt{\varepsilon}\leq \sum_{l} W^{out}_{1l} relu(V_l^{\mu_k}-\theta_l) \leq \sqrt{\varepsilon} \tag{D4} \\
 \rightarrow W^{out}_{1q} \approx \mathcal{O}(\sqrt{\varepsilon}), \; \forall q,k \; | \; V^{\mu_k}_q>\theta_q \tag{D5}\\
 1-\sqrt{\varepsilon}\leq \sum_{l} W^{out}_{1l} relu(V_l^{\mu_1}-\theta_l) \leq 1+\sqrt{\varepsilon} \tag{D6}\\
 \rightarrow W^{out}_{1m} \gg \mathcal{O}(\sqrt{\varepsilon}), \; \forall m \; | \; V^{\mu_1}_m>\theta_m  \tag{D7}
\end{align}
the conditions above can be satisfied only when the indexes $q$ do not completely overlap with the indexes $m$, i.e. the representations do not totally overlap and a vector of thresholds is introduced to separate the ensemble of nodes that are active for different classes. We note that this simple proof holds in the case of positive output weights and positive reservoir activities only.

\subsection{Interpretation of the optimal fifty percent of active nodes for memory capacity}

The aim of the task faced in section 2.2 (Main text) is to measure the memory capacity of the model and the stability of the solution found by the model. The results show a robust optimal level of sparsity of $0.5$ despite the specific values of noise level and number of output classes. Such level of sparsity maximises the probability that different representations have at least one node that is not in common. Given $N$ nodes and an undefined input $s_i$, the ensemble of active nodes for that signal can be imagined as a random sample of $p \times N$ nodes from the total possible ensemble of neurons. Thus, each representation can be imagined as the extraction of $p \times N$ elements from an urn of $N$ elements, where $p$ is the imposed percentage of active nodes. In order to have representations that do not completely overlap we need to maximise the number of possible outcomes of the extraction, and this will guarantee that at least one node is different among the various representations. The number of possible extractions of active nodes   
corresponds to the number of combinations without repetitions:
\begin{equation}
\tilde{N}=\dfrac{N!}{(N-p\times N)!(p\times N)!} \tag{D8}
\end{equation}   
which has a maximum at $p=0.5$.
Thus, $p=0.5$ is the sparsity level that maximises the probability that, given an undefined ensemble of input stimuli, the corresponding representations will have at least one non-overlapping neuron.

\bibliographystyle{unsrt}
\bibliography{Manneschi_Lin_Vsilaki}

\begin{thebibliography}{10}
\providecommand{\url}[1]{#1}
\csname url@samestyle\endcsname
\providecommand{\newblock}{\relax}
\providecommand{\bibinfo}[2]{#2}
\providecommand{\BIBentrySTDinterwordspacing}{\spaceskip=0pt\relax}
\providecommand{\BIBentryALTinterwordstretchfactor}{4}
\providecommand{\BIBentryALTinterwordspacing}{\spaceskip=\fontdimen2\font plus
\BIBentryALTinterwordstretchfactor\fontdimen3\font minus
  \fontdimen4\font\relax}
\providecommand{\BIBforeignlanguage}[2]{{%
\expandafter\ifx\csname l@#1\endcsname\relax
\typeout{** WARNING: IEEEtran.bst: No hyphenation pattern has been}%
\typeout{** loaded for the language `#1'. Using the pattern for}%
\typeout{** the default language instead.}%
\else
\language=\csname l@#1\endcsname
\fi
#2}}
\providecommand{\BIBdecl}{\relax}
\BIBdecl

\bibitem{tsodyks1988enhanced}
M.~V. Tsodyks and M.~V. Feigel'man, ``The enhanced storage capacity in neural
  networks with low activity level,'' \emph{EPL (Europhysics Letters)}, vol.~6,
  no.~2, p. 101, 1988.

\bibitem{tsodyks1988associative}
M.~Tsodyks, ``Associative memory in asymmetric diluted network with low level
  of activity,'' \emph{EPL (Europhysics Letters)}, vol.~7, no.~3, p. 203, 1988.

\bibitem{derrida1987exactly}
B.~Derrida, E.~Gardner, and A.~Zippelius, ``An exactly solvable asymmetric
  neural network model,'' \emph{EPL (Europhysics Letters)}, vol.~4, no.~2, p.
  167, 1987.

\bibitem{amit1985storing}
D.~J. Amit, H.~Gutfreund, and H.~Sompolinsky, ``Storing infinite numbers of
  patterns in a spin-glass model of neural networks,'' \emph{Physical Review
  Letters}, vol.~55, no.~14, p. 1530, 1985.

\bibitem{romani2013scaling}
S.~Romani, I.~Pinkoviezky, A.~Rubin, and M.~Tsodyks, ``Scaling laws of
  associative memory retrieval,'' \emph{Neural computation}, vol.~25, no.~10,
  pp. 2523--2544, 2013.

\bibitem{hastie2015statistical}
T.~Hastie, R.~Tibshirani, and M.~Wainwright, \emph{Statistical learning with
  sparsity: the lasso and generalizations}.\hskip 1em plus 0.5em minus
  0.4em\relax Chapman and Hall/CRC, 2015.

\bibitem{NIPS2016_6504}
\BIBentryALTinterwordspacing
W.~Wen, C.~Wu, Y.~Wang, Y.~Chen, and H.~Li, ``Learning structured sparsity in
  deep neural networks,'' in \emph{Advances in Neural Information Processing
  Systems 29}, D.~D. Lee, M.~Sugiyama, U.~V. Luxburg, I.~Guyon, and R.~Garnett,
  Eds.\hskip 1em plus 0.5em minus 0.4em\relax Curran Associates, Inc., 2016,
  pp. 2074--2082. [Online]. Available:
  \url{http://papers.nips.cc/paper/6504-learning-structured-sparsity-in-deep-neural-networks.pdf}
\BIBentrySTDinterwordspacing

\bibitem{srivastava2014dropout}
N.~Srivastava, G.~Hinton, A.~Krizhevsky, I.~Sutskever, and R.~Salakhutdinov,
  ``Dropout: a simple way to prevent neural networks from overfitting,''
  \emph{The journal of machine learning research}, vol.~15, no.~1, pp.
  1929--1958, 2014.

\bibitem{rasmussen2012model}
P.~M. Rasmussen, L.~K. Hansen, K.~H. Madsen, N.~W. Churchill, and S.~C.
  Strother, ``Model sparsity and brain pattern interpretation of classification
  models in neuroimaging,'' \emph{Pattern Recognition}, vol.~45, no.~6, pp.
  2085--2100, 2012.

\bibitem{rolls1995sparseness}
E.~T. Rolls and M.~J. Tovee, ``Sparseness of the neuronal representation of
  stimuli in the primate temporal visual cortex,'' \emph{Journal of
  neurophysiology}, vol.~73, no.~2, pp. 713--726, 1995.

\bibitem{honegger2011cellular}
K.~S. Honegger, R.~A. Campbell, and G.~C. Turner, ``Cellular-resolution
  population imaging reveals robust sparse coding in the drosophila mushroom
  body,'' \emph{Journal of Neuroscience}, vol.~31, no.~33, pp.
  11\,772--11\,785, 2011.

\bibitem{lin2014sparse}
A.~C. Lin, A.~M. Bygrave, A.~De~Calignon, T.~Lee, and G.~Miesenb{\"o}ck,
  ``Sparse, decorrelated odor coding in the mushroom body enhances learned odor
  discrimination,'' \emph{Nature neuroscience}, vol.~17, no.~4, p. 559, 2014.

\bibitem{turner2008olfactory}
G.~C. Turner, M.~Bazhenov, and G.~Laurent, ``Olfactory representations by
  drosophila mushroom body neurons,'' \emph{Journal of neurophysiology},
  vol.~99, no.~2, pp. 734--746, 2008.

\bibitem{gruntman2013integration}
E.~Gruntman and G.~C. Turner, ``Integration of the olfactory code across
  dendritic claws of single mushroom body neurons,'' \emph{Nature
  neuroscience}, vol.~16, no.~12, p. 1821, 2013.

\bibitem{li2013transformation}
H.~Li, Y.~Li, Z.~Lei, K.~Wang, and A.~Guo, ``Transformation of odor selectivity
  from projection neurons to single mushroom body neurons mapped with
  dual-color calcium imaging,'' \emph{Proceedings of the National Academy of
  Sciences}, vol. 110, no.~29, pp. 12\,084--12\,089, 2013.

\bibitem{perez2002oscillations}
J.~Perez-Orive, O.~Mazor, G.~C. Turner, S.~Cassenaer, R.~I. Wilson, and
  G.~Laurent, ``Oscillations and sparsening of odor representations in the
  mushroom body,'' \emph{Science}, vol. 297, no. 5580, pp. 359--365, 2002.

\bibitem{jeanne2015convergence}
J.~M. Jeanne and R.~I. Wilson, ``Convergence, divergence, and reconvergence in
  a feedforward network improves neural speed and accuracy,'' \emph{Neuron},
  vol.~88, no.~5, pp. 1014--1026, 2015.

\bibitem{azouz2000dynamic}
R.~Azouz and C.~M. Gray, ``Dynamic spike threshold reveals a mechanism for
  synaptic coincidence detection in cortical neurons in vivo,''
  \emph{Proceedings of the National Academy of Sciences}, vol.~97, no.~14, pp.
  8110--8115, 2000.

\bibitem{grubb2010activity}
M.~S. Grubb and J.~Burrone, ``Activity-dependent relocation of the axon initial
  segment fine-tunes neuronal excitability,'' \emph{Nature}, vol. 465, no.
  7301, p. 1070, 2010.

\bibitem{du2017reservoir}
C.~Du, F.~Cai, M.~A. Zidan, W.~Ma, S.~H. Lee, and W.~D. Lu, ``Reservoir
  computing using dynamic memristors for temporal information processing,''
  \emph{Nature communications}, vol.~8, no.~1, pp. 1--10, 2017.

\bibitem{kulkarni2012memristor}
M.~S. Kulkarni and C.~Teuscher, ``Memristor-based reservoir computing,'' in
  \emph{2012 IEEE/ACM international symposium on nanoscale architectures
  (NANOARCH)}.\hskip 1em plus 0.5em minus 0.4em\relax IEEE, 2012, pp. 226--232.

\bibitem{zhu2020memristor}
X.~Zhu, Q.~Wang, and W.~D. Lu, ``Memristor networks for real-time neural
  activity analysis,'' \emph{Nature communications}, vol.~11, no.~1, pp. 1--9,
  2020.

\bibitem{vandoorne2014experimental}
K.~Vandoorne, P.~Mechet, T.~Van~Vaerenbergh, M.~Fiers, G.~Morthier,
  D.~Verstraeten, B.~Schrauwen, J.~Dambre, and P.~Bienstman, ``Experimental
  demonstration of reservoir computing on a silicon photonics chip,''
  \emph{Nature communications}, vol.~5, no.~1, pp. 1--6, 2014.

\bibitem{paquot2012optoelectronic}
Y.~Paquot, F.~Duport, A.~Smerieri, J.~Dambre, B.~Schrauwen, M.~Haelterman, and
  S.~Massar, ``Optoelectronic reservoir computing,'' \emph{Scientific reports},
  vol.~2, p. 287, 2012.

\bibitem{nakajima2020physical}
K.~Nakajima, ``Physical reservoir computing—an introductory perspective,''
  \emph{Japanese Journal of Applied Physics}, vol.~59, no.~6, p. 060501, 2020.

\bibitem{jaeger2007optimization}
H.~Jaeger, M.~Luko{\v{s}}evi{\v{c}}ius, D.~Popovici, and U.~Siewert,
  ``Optimization and applications of echo state networks with leaky-integrator
  neurons,'' \emph{Neural networks}, vol.~20, no.~3, pp. 335--352, 2007.

\bibitem{rodan2010minimum}
A.~Rodan and P.~Tino, ``Minimum complexity echo state network,'' \emph{IEEE
  transactions on neural networks}, vol.~22, no.~1, pp. 131--144, 2010.

\bibitem{cui2012architecture}
H.~Cui, X.~Liu, and L.~Li, ``The architecture of dynamic reservoir in the echo
  state network,'' \emph{Chaos: An Interdisciplinary Journal of Nonlinear
  Science}, vol.~22, no.~3, p. 033127, 2012.

\bibitem{gallicchio2017deep}
C.~Gallicchio, A.~Micheli, and L.~Pedrelli, ``Deep reservoir computing: A
  critical experimental analysis,'' \emph{Neurocomputing}, vol. 268, pp.
  87--99, 2017.

\bibitem{manneschi2020}
L.~Manneschi, M.~O.~A. Ellis, G.~Gigante, A.~C. Lin, P.~Del~Giudice, and
  E.~Vasilaki, ``Exploiting multiple timescales in hierarchical echo state
  networks,'' \emph{Frontiers}, 2021.

\bibitem{bianchi2020reservoir}
F.~M. Bianchi, S.~Scardapane, S.~L{\o}kse, and R.~Jenssen, ``Reservoir
  computing approaches for representation and classification of multivariate
  time series,'' \emph{IEEE Transactions on Neural Networks and Learning
  Systems}, 2020.

\bibitem{huang2011learning}
J.~Huang, T.~Zhang, and D.~Metaxas, ``Learning with structured sparsity,''
  \emph{Journal of Machine Learning Research}, vol.~12, no. Nov, pp.
  3371--3412, 2011.

\bibitem{candes2008enhancing}
E.~J. Candes, M.~B. Wakin, and S.~P. Boyd, ``Enhancing sparsity by reweighted l
  1 minimization,'' \emph{Journal of Fourier analysis and applications},
  vol.~14, no. 5-6, pp. 877--905, 2008.

\bibitem{lukovsevivcius2012practical}
M.~Luko{\v{s}}evi{\v{c}}ius, ``A practical guide to applying echo state
  networks,'' in \emph{Neural networks: Tricks of the trade}.\hskip 1em plus
  0.5em minus 0.4em\relax Springer, 2012, pp. 659--686.

\bibitem{ma2016functional}
Q.~Ma, L.~Shen, W.~Chen, J.~Wang, J.~Wei, and Z.~Yu, ``Functional echo state
  network for time series classification,'' \emph{Information Sciences}, vol.
  373, pp. 1--20, 2016.

\bibitem{schaetti2016echo}
N.~Schaetti, M.~Salomon, and R.~Couturier, ``Echo state networks-based
  reservoir computing for mnist handwritten digits recognition,'' in \emph{2016
  IEEE Intl Conference on Computational Science and Engineering (CSE) and IEEE
  Intl Conference on Embedded and Ubiquitous Computing (EUC) and 15th Intl
  Symposium on Distributed Computing and Applications for Business Engineering
  (DCABES)}.\hskip 1em plus 0.5em minus 0.4em\relax IEEE, 2016, pp. 484--491.

\bibitem{torrejon2017neuromorphic}
J.~Torrejon, M.~Riou, F.~A. Araujo, S.~Tsunegi, G.~Khalsa, D.~Querlioz,
  P.~Bortolotti, V.~Cros, K.~Yakushiji, A.~Fukushima \emph{et~al.},
  ``Neuromorphic computing with nanoscale spintronic oscillators,''
  \emph{Nature}, vol. 547, no. 7664, pp. 428--431, 2017.

\bibitem{tanaka2019recent}
G.~Tanaka, T.~Yamane, J.~B. H{\'e}roux, R.~Nakane, N.~Kanazawa, S.~Takeda,
  H.~Numata, D.~Nakano, and A.~Hirose, ``Recent advances in physical reservoir
  computing: A review,'' \emph{Neural Networks}, vol. 115, pp. 100--123, 2019.

\bibitem{chandar2019towards}
S.~Chandar, C.~Sankar, E.~Vorontsov, S.~E. Kahou, and Y.~Bengio, ``Towards
  non-saturating recurrent units for modelling long-term dependencies,'' in
  \emph{Proceedings of the AAAI Conference on Artificial Intelligence},
  vol.~33, 2019, pp. 3280--3287.

\bibitem{kemker2018measuring}
R.~Kemker, M.~McClure, A.~Abitino, T.~L. Hayes, and C.~Kanan, ``Measuring
  catastrophic forgetting in neural networks,'' in \emph{Thirty-second AAAI
  conference on artificial intelligence}, 2018.

\bibitem{serra2018overcoming}
J.~Serra, D.~Suris, M.~Miron, and A.~Karatzoglou, ``Overcoming catastrophic
  forgetting with hard attention to the task,'' \emph{arXiv preprint
  arXiv:1801.01423}, 2018.

\bibitem{hallem2006coding}
E.~A. Hallem and J.~R. Carlson, ``Coding of odors by a receptor repertoire,''
  \emph{Cell}, vol. 125, no.~1, pp. 143--160, 2006.

\bibitem{olsen2010divisive}
S.~R. Olsen, V.~Bhandawat, and R.~I. Wilson, ``Divisive normalization in
  olfactory population codes,'' \emph{Neuron}, vol.~66, no.~2, pp. 287--299,
  2010.

\bibitem{luo2010generating}
S.~X. Luo, R.~Axel, and L.~Abbott, ``Generating sparse and selective
  third-order responses in the olfactory system of the fly,'' \emph{Proceedings
  of the National Academy of Sciences}, vol. 107, no.~23, pp. 10\,713--10\,718,
  2010.

\bibitem{parnas2013odor}
M.~Parnas, A.~C. Lin, W.~Huetteroth, and G.~Miesenb{\"o}ck, ``Odor
  discrimination in drosophila: from neural population codes to behavior,''
  \emph{Neuron}, vol.~79, no.~5, pp. 932--944, 2013.

\bibitem{krishnamurthy2017disorder}
K.~Krishnamurthy, A.~M. Hermundstad, T.~Mora, A.~M. Walczak, and
  V.~Balasubramanian, ``Disorder and the neural representation of complex
  odors: smelling in the real world,'' \emph{arXiv preprint arXiv:1707.01962},
  2017.

\bibitem{caron2013random}
S.~J. Caron, V.~Ruta, L.~Abbott, and R.~Axel, ``Random convergence of olfactory
  inputs in the drosophila mushroom body,'' \emph{Nature}, vol. 497, no. 7447,
  p. 113, 2013.

\bibitem{song2005highly}
S.~Song, P.~J. Sj{\"o}str{\"o}m, M.~Reigl, S.~Nelson, and D.~B. Chklovskii,
  ``{Highly nonrandom features of synaptic connectivity in local cortical
  circuits.}'' \emph{PLoS biology}, vol.~3, no.~3, p. e68, Mar. 2005.

\bibitem{kingma2014adam}
D.~P. Kingma and J.~Ba, ``Adam: A method for stochastic optimization,''
  \emph{arXiv preprint arXiv:1412.6980}, 2014.

\bibitem{deng2012mnist}
L.~Deng, ``The mnist database of handwritten digit images for machine learning
  research [best of the web],'' \emph{IEEE Signal Processing Magazine},
  vol.~29, no.~6, pp. 141--142, 2012.

\bibitem{arora2019does}
G.~Arora, A.~Rahimi, and T.~Baldwin, ``Does an lstm forget more than a cnn? an
  empirical study of catastrophic forgetting in nlp,'' in \emph{Proceedings of
  the The 17th Annual Workshop of the Australasian Language Technology
  Association}, 2019, pp. 77--86.

\bibitem{zheng2018complete}
Z.~Zheng, J.~S. Lauritzen, E.~Perlman, C.~G. Robinson, M.~Nichols, D.~Milkie,
  O.~Torrens, J.~Price, C.~B. Fisher, N.~Sharifi, S.~A. Calle-Schuler,
  L.~Kmecova, I.~J. Ali, B.~Karsh, E.~T. Trautman, J.~A. Bogovic,
  P.~Hanslovsky, G.~S. X.~E. Jefferis, M.~Kazhdan, K.~Khairy, S.~Saalfeld,
  R.~D. Fetter, and D.~D. Bock, ``{A Complete Electron Microscopy Volume of the
  Brain of Adult Drosophila melanogaster.}'' \emph{Cell}, vol. 174, no.~3, pp.
  730--743.e22, Jul. 2018.

\bibitem{hermans2012recurrent}
M.~Hermans and B.~Schrauwen, ``Recurrent kernel machines: Computing with
  infinite echo state networks,'' \emph{Neural Computation}, vol.~24, no.~1,
  pp. 104--133, 2012.

\bibitem{markovic2019reservoir}
D.~Markovi{\'c}, N.~Leroux, M.~Riou, F.~Abreu~Araujo, J.~Torrejon, D.~Querlioz,
  A.~Fukushima, S.~Yuasa, J.~Trastoy, P.~Bortolotti \emph{et~al.}, ``Reservoir
  computing with the frequency, phase, and amplitude of spin-torque
  nano-oscillators,'' \emph{Applied Physics Letters}, vol. 114, no.~1, p.
  012409, 2019.

\bibitem{romera2018vowel}
M.~Romera, P.~Talatchian, S.~Tsunegi, F.~A. Araujo, V.~Cros, P.~Bortolotti,
  J.~Trastoy, K.~Yakushiji, A.~Fukushima, H.~Kubota \emph{et~al.}, ``Vowel
  recognition with four coupled spin-torque nano-oscillators,'' \emph{Nature},
  vol. 563, no. 7730, p. 230, 2018.

\end{thebibliography}

\end{document}